\documentclass[11pt]{article}

\usepackage[preprint]{acl}

\usepackage{times}
\usepackage{latexsym}
\usepackage{tabularx}

\usepackage[T1]{fontenc}

\usepackage[utf8]{inputenc}

\usepackage{microtype}

\usepackage{inconsolata}

\usepackage{graphicx}
\usepackage{booktabs}

\usepackage{amsmath}
\usepackage{multirow}
\usepackage{float}
\usepackage{enumitem}
\usepackage{tcolorbox}

\title{Does Reasoning Improve Psychological Depth in Large Language Models?}

\author{%
  \textbf{Ruichen Zheng}\textnormal{\textsuperscript{1}}\thanks{Now at Arizona State University. Correspondence: \texttt{rzheng46@asu.edu}.} \quad
  \textbf{Yihe Wang}\textnormal{\textsuperscript{2}}\thanks{Work partly done while visiting UCLA.} \quad
  \textbf{Fabrice Harel-Canada}\textnormal{\textsuperscript{1,3}} \quad
  \textbf{Sara Khosravi}\textnormal{\textsuperscript{1}} \\
  \textbf{Zeynep Senahan Yildiz}\textnormal{\textsuperscript{1}} \quad
  \textbf{Amit Sahai}\textnormal{\textsuperscript{1}} \quad
  \textbf{Nanyun Peng}\textnormal{\textsuperscript{1,4}} \\
  \vspace{0.2em} \\
  \textsuperscript{1}University of California, Los Angeles \quad
  \textsuperscript{2}Cornell University \\
  \textsuperscript{3}LA General Hospital \quad
  \textsuperscript{4}Google \\
}

\begin{document}
\maketitle

\begin{abstract}
Benchmark gains from reasoning-enhanced large language models encourage the assumption that these models should broadly perform better on complex generative tasks, but this remains underexplored in open-ended, creative domains shaped by nuances of reader experience. We tested this assumption through the lens of psychological depth in short stories. We compared DeepSeek-R1 vs.\ DeepSeek-V3, our closest available contrast for reasoning-oriented post-training, and GPT-5 vs.\ GPT-4o, through 60 blinded pairwise evaluations in which seven human annotators and an ensemble LLM-as-a-Judge rated prompt-matched story outputs. Human evaluations showed no universal reasoning advantage: GPT-5 was slightly preferred over GPT-4o (60.0--62.9\% across two reasoning-effort settings), whereas DeepSeek-R1 fell behind DeepSeek-V3 (42.9\%). Although this pattern was marked by low inter-annotator agreement (Krippendorff's $\alpha = 0.070$), our annotation-quality checks suggested genuine evaluative subjectivity, plausibly reflecting distinct annotator decision schemes rather than annotation failure. In contrast, the LLM-as-a-Judge showed a strong pro-reasoning bias (98.3\% of pairs vs. 55.2\% of human votes), consistent across its five configurations and partly tracking sentence length. These findings suggest that, at the frontier, reasoning-related quality differences are small enough to be dominated by reader subjectivity, and that LLM-as-a-Judge evaluators can obscure this subjectivity in nuanced creative-generation tasks.

\end{abstract}

\section{Introduction}
\label{sec:intro}

Reasoning-enhanced language models are routinely evaluated on mathematical, coding, and factual benchmarks \citep{openai-o1,snell2024scaling, chen2026learning, li202512surveyreasoning, zeng2025thinkingloudreasoningmodels,yu2025benchmarking, 10.1145/3774896}. Models such as GPT-5 and DeepSeek-R1 undergo post-training aimed at strengthening deliberative reasoning, often through reinforcement learning or related reasoning-oriented methods \citep{singh2025openaigpt5card, Guo_2025}. However, much less is known about how such post-training affects creative generation. In particular, performance gains on standard benchmarks invite the assumption that the capabilities behind these gains should also generalize to complex, open-ended tasks such as fiction writing, though this has yet to be investigated directly.

We investigated this question through the lens of \emph{psychological depth} in short stories. We constructed 60 blinded pairwise story comparisons across three frontier model matchups, namely GPT-5 versus GPT-4o \citep{hurst2024gpt} under two reasoning effort settings and DeepSeek-R1 versus DeepSeek-V3. To evaluate these pairs, we used the Psychological Depth Scale (PDS) \citep{harelcanada2024measuringpsychologicaldepthlanguage}, which measures authenticity, empathy, engagement, emotion provocation, and narrative complexity. Each story pair was evaluated by seven human annotators and an LLM-as-a-Judge ensemble. We also collected an overall preference to examine how preferences on individual PDS dimensions relate to readers’ overall judgments. The matchups play complementary roles: DeepSeek-R1 versus DeepSeek-V3 is our closest available, though not clean, contrast for reasoning-oriented post-training, while GPT-5 versus GPT-4o is a model-level comparison that brings controlled human evaluation to a widely discussed transition (``GPT-5 reads colder than GPT-4o'') so far characterized mostly by anecdote.

This setup allowed us to test not only whether reasoning-enhanced large language models are preferred, but also whether such preferences hold across model families and readers. More specifically, we asked three questions: \textbf{RQ1.} Do humans prefer reasoning-enhanced LLMs over their non-reasoning counterparts on storywriting tasks? \textbf{RQ2.} Are these preferences consistent across annotators and psychological-depth dimensions? \textbf{RQ3.} Do LLM-as-Judge evaluators recover the same preference patterns as human annotators?

\begin{figure*}[t]
    \centering
    \includegraphics[width=0.7\textwidth]{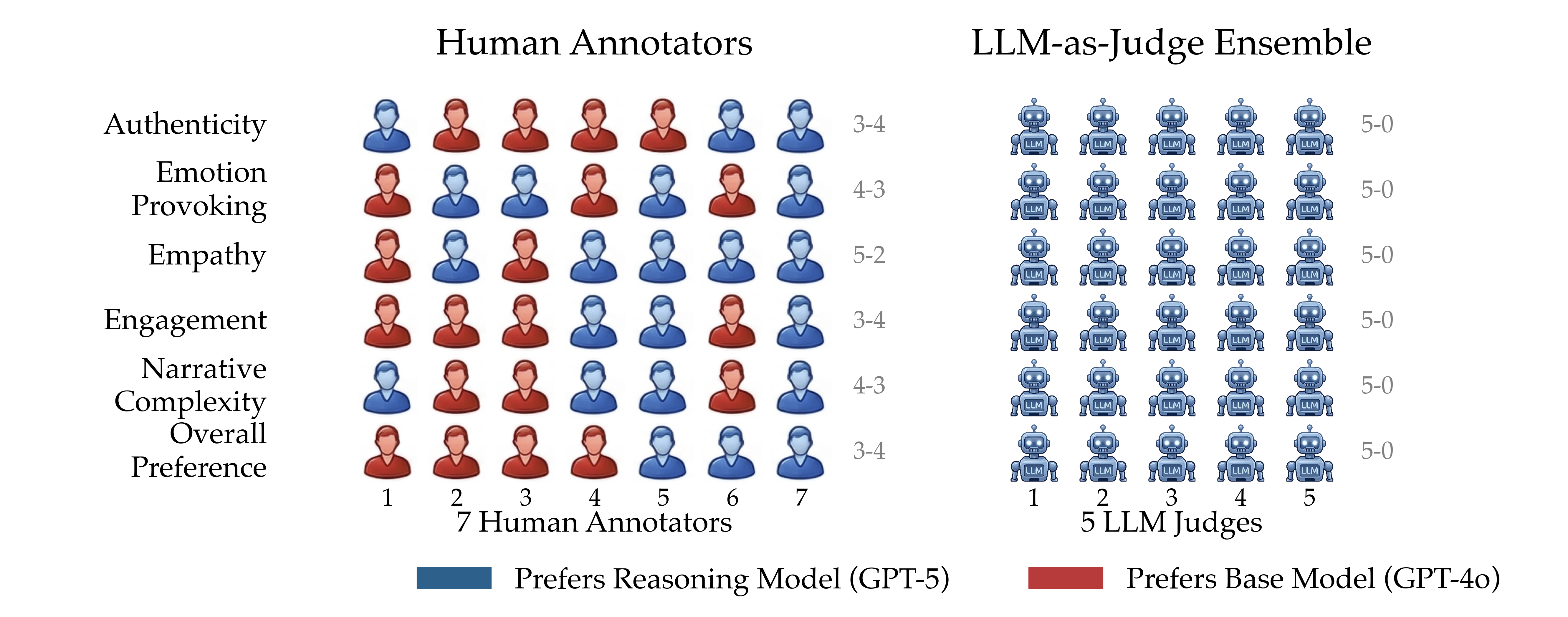}%
    \caption{\textbf{Evaluating Psychological Depth: Human Subjectivity vs. LLM Consensus.} Evaluation patterns across five psychological dimensions for a representative story pair (Pair 24, GPT-5 vs.\ GPT-4o). (\textbf{Left}) Human annotators displayed diverse preferences, reflecting subjective interpretations. (\textbf{Right}) All five LLM-as-a-Judge configurations in the ensemble collapsed this subjectivity into unanimous preference for the reasoning model.
    }
\label{fig:teaser}
\end{figure*}

Taken together, our findings suggest a more heterogeneous picture than standard benchmark gains alone would lead one to expect. In comparisons evaluated by human annotators, GPT-5 was slightly preferred over GPT-4o, whereas DeepSeek-R1 fell behind DeepSeek-V3, though neither difference was significant. This contrasts with earlier PDS evaluations, where quality gaps between stories were large enough to yield more reliable ratings; at the frontier, any quality difference attributable to reasoning was small enough to be dominated by structured reader subjectivity rather than to produce annotator consensus. The human evaluations also showed substantial disagreement, and our quality checks suggest that this reflects genuine evaluative subjectivity rather than annotation failure, plausibly because annotators rely on distinct rationales and weight psychological-depth dimensions differently. In contrast, despite being specifically optimized to maximize agreement with human annotations on the original PDS dataset ($\rho$ = 0.644), the LLM-as-a-Judge evaluator remained strongly pro-reasoning across both model families, which may obscure meaningful variation visible in the human results. Figure~\ref{fig:teaser} illustrates this contrast for a representative story pair: human annotators displayed diverse preferences across PDS dimensions, whereas the LLM-as-Judge ensemble collapsed into unanimous agreement.

Our paper makes three contributions. First, we present a controlled human evaluation of frontier reasoning and non-reasoning models on short-story generation through the lens of psychological depth. Second, we provide evidence that annotator disagreement is a meaningful part of the evaluative phenomenon itself rather than random noise, and we relate that disagreement to distinct dimension-weighting schemes across annotators. Third, we show that a validated LLM-as-a-Judge evaluator has a strong pro-reasoning bias, preferring the reasoning model in 98.3\% of pairs versus 55.2\% of human votes, consistently across judge configurations and partly tracking surface features, posing a challenge for nuanced story evaluation and for alignment with human preferences.
\section{Related Work}
\label{sec:related}

\paragraph{Psychological Depth and Creativity Evaluation.} Evaluating creative text remains a longstanding challenge in NLP \citep{gomez-rodriguez-williams-2023-confederacy}. Traditional metrics such as perplexity often capture fluency rather than the kinds of qualities that matter in literary evaluation \citep{lu2025rethinking}, and recent work has questioned whether LLMs exhibit genuine creativity or merely reproduce familiar surface patterns \citep{chakrabarty2024art}. Most relevant to our study, \citet{harelcanada2024measuringpsychologicaldepthlanguage} introduce the Psychological Depth Scale (PDS), a five-dimensional framework for evaluating authenticity, empathy, engagement, emotion provocation, and narrative complexity (abbreviated as AUT, EMP, ENG, PROV, and NCOM) in fictional stories, and show that story ratings averaged across annotators are moderately reliable (Intraclass Correlation Coefficient (ICC) = 0.60). We build on that framework in a more demanding frontier setting, where quality differences between model outputs are narrower and pairwise comparison better fits the task \citep{liu2024aligning}. Importantly, research on annotator disagreement treats variance in subjective annotation not as noise but as a meaningful signal of value pluralism and task ambiguity \citep{clark-etal-2021-thats, sandri-etal-2023-dont, kirk2024prism}. Our analysis extends this perspective by relating disagreement to differences in how annotators weight dimensions of psychological depth.

\paragraph{Reasoning-Enhanced LLMs Beyond Benchmarks.}
Test-time compute scaling has emerged as a powerful paradigm for improving LLM capabilities. OpenAI's o1 \citep{openai-o1} and GPT-5 \citep{singh2025openaigpt5card} apply extended reasoning chains during inference, while DeepSeek-R1 \citep{Guo_2025} achieves similar effects through reinforcement learning. \citet{snell2024scaling} demonstrate that scaling inference-time compute can outperform parameter scaling on reasoning benchmarks. However, whether these gains transfer to subjective generative tasks remains an open question. DeepSeek-R1 is initialized from DeepSeek-V3-Base, with post-training organized around reasoning-oriented RL and supporting reasoning-data SFT stages \citep{Guo_2025, liu2024deepseek}; its comparison with DeepSeek-V3 is the closest publicly documented contrast for studying how reasoning post-training affects narrative quality, though not a clean ablation. Anecdotal concerns and widespread public sentiment suggest that reasoning-oriented post-training may alter writing style or creativity, but systematic evaluation on narrative and psychological dimensions is lacking.

\paragraph{LLM-as-a-Judge Evaluation.} The LLM-as-a-Judge paradigm has become a standard tool for scalable assessment \citep{zheng2023judgingllmasajudgemtbenchchatbot}. Prior work suggests that pairwise formats can align well with human judgment \citep{liu2024aligning}, though more recent work shows that pairwise comparison can also amplify evaluator biases \citep{jeong-etal-2025-comparative}. These concerns are especially relevant for our setting, where the evaluation target is subjective and frontier outputs are often close in quality. They also motivate our decision to ground the evaluator in human-validated scalar judgments and derive pairwise preferences post hoc.

\section{\textsc{PsychDepth Reasoning} Dataset}
\label{sec:dataset}

To examine how reasoning-oriented post-training affects creative writing, we constructed \textsc{PsychDepth Reasoning}. This dataset contains 120 stories organized into 60 prompt-matched pairs from 20 carefully curated premises.

\paragraph{Defining ``Reasoning-Enhanced.''}
In this paper, we use \textit{reasoning-enhanced} to refer to models whose post-training emphasizes deliberative reasoning behavior, typically via reinforcement learning or related reasoning-oriented optimization. In our study, this includes DeepSeek-R1, which uses reasoning-oriented RL \citep{Guo_2025}, and GPT-5, which is trained to reason with extended chain-of-thought before answering \citep{singh2025openaigpt5card}. We use \textit{base} model to refer to counterparts without such reasoning-oriented post-training, i.e. DeepSeek-V3 and GPT-4o. This definition concerns weight-level post-training, not test-time reasoning length (see Limitations).

\paragraph{Premise Selection.}
To elicit psychological depth, we curated 20 premises that provide contextual framing for character development while leaving the narrative direction open. We sourced 15 premises from Reddit’s r/WritingPrompts\footnote{\url{https://www.reddit.com/r/WritingPrompts}}, a highly popular online community known for its thematic diversity and structured, high-concept story prompts, and 5 premises from Reedsy\footnote{\url{https://reedsy.com/creative-writing-prompts/}}, a professional platform offering traditional literary prompts. To prevent data contamination, we restricted Reddit selection to prompts posted after the known training-data cutoff dates for GPT-5 and DeepSeek-R1, then ranked candidates by all-time upvotes and manually filtered for quality. The full list of all 20 premises appears in Appendix~\ref{app:premises}.

\paragraph{Model Pairings and Pair Construction.}
We selected the GPT and DeepSeek model families to compare prompt-matched story outputs from reasoning and non-reasoning models. For each premise, we paired one reasoning-model story with one base-model story from the same family. All stories were generated with a common zero-shot story-writing template, with only limited family-specific wording adjustments for word-count compliance (identical within each pair). Full prompts and sampling settings appear in Appendix~\ref{app:generation}, and surface statistics by model in Appendix~\ref{app:surface}.

The two families play complementary roles. \textbf{DeepSeek-R1} and \textbf{DeepSeek-V3} share the DeepSeek-V3-Base model and diverge mainly through reasoning-oriented post-training, making them our most controlled available anchor for reasoning claims, though not a clean ablation (20 pairs). For the GPT family, we evaluated \textbf{GPT-5}, one of the most widely used frontier reasoning models, against \textbf{GPT-4o} under two distinct reasoning budget settings, defined as the number of reasoning tokens generated prior to the final response: \textbf{GPT-5 (auto)}, where the model autonomously determines the reasoning length ($7{,}846 \pm 3{,}117$ reasoning tokens on average), and \textbf{GPT-5 (high)} ($11{,}606 \pm 3{,}557$ reasoning tokens, roughly $1.5 \times$ the auto budget), with 20 pairs each, to test whether increased reasoning compute affects perceived psychological depth. Because GPT-5 and GPT-4o likely also differ in data, architecture, and alignment, we treat this pairing as an ecologically valid complement: evidence about reasoning as users encounter it, but weaker evidence about reasoning post-training in isolation.

\section{Human Study}
\label{sec:human}
We adapted the \citet{harelcanada2024measuringpsychologicaldepthlanguage} framework to a blinded pairwise protocol. The Institutional Review Board (IRB) deemed this study exempt.

\subsection{Annotator Recruitment and Training}
\label{sec:training}

We recruited undergraduate students from the English and Psychology departments at a large research university using targeted interest forms. From 70 applicants, we selected the 7 most promising candidates based on English proficiency, interest in the research, and experience reviewing short stories, with cohort size determined by budget constraints.

We conducted comprehensive online onboarding sessions covering the five PDS metrics and our custom Label Studio annotation interface \citep{LabelStudio}. The participants then independently completed the annotations in seven days and received \$100 each.

\subsection{Annotation Protocol}
\label{sec:protocol}

For each of the 60 pairs, two stories were presented side-by-side in a randomized order, with the PDS rubric definitions shown at the top of the page. For each pair, annotators provided six forced-choice pairwise preference judgments (with no tie option): one for each of the five PDS dimensions (authenticity, empathy, engagement, emotion provocation, and narrative complexity) and one overall preference judgment (PREF; see Appendix~\ref{app:interface} on annotation interface). Annotators could also submit free-text justifications. This produced $7 \times 60 \times 6 = 2{,}520$ pairwise preference judgments, along with 112 free-text justifications. To monitor annotation quality, we logged per-pair completion times and conducted a Leave-One-Annotator-Out (LOAO) stability analysis and administered a post-study survey; all quality checks are reported in Appendix~\ref{app:annotator_time} and~\ref{app:post_survey}.

\paragraph{Statistical analysis.}\label{sec:stats} Because judgments are clustered by story pair and by annotator, we report observed preference rates with 95\% intervals from an intercept-only logistic generalized linear mixed model (GLMM) with crossed random intercepts for pair and annotator, fit per matchup and dimension by maximum likelihood with profile-likelihood intervals (Appendix~\ref{app:stats}); intervals are pointwise and unadjusted for multiplicity. Annotator variance exceeds pair variance in 12 of 18 fits (medians 0.38 vs.\ 0.28). Pair- and annotator-level cluster bootstraps serve as sensitivity checks (Appendix~\ref{app:stats}).

\section{LLM-as-a-Judge Evaluator}
\label{sec:auto}

To compare automatic and human judgments, we constructed an LLM-as-a-Judge pipeline that extends a zero-shot evaluation framework to a controlled few-shot paradigm.

\subsection{Evaluator Construction}

Our pipeline uses DSPy \citep{dspy} to ensure that LLM judges return 1-to-5 Likert ratings and retains the Mixture-of-Personas prompting strategy of the original PDS study, which primes evaluators with distinct analytical perspectives \citep{salewski2023context}. We partitioned the original 97-story dataset and its human annotations \citep{harelcanada2024measuringpsychologicaldepthlanguage} into a 20\% demonstration pool and an 80\% validation split, and selected representative few-shot examples from the pool by $k$-means clustering of PDS score vectors, drawing on prior selective-annotation work that emphasizes diverse, representative examples for in-context learning \citep{su2022selective}.

We used a scalar-rating evaluator rather than a directly pairwise-trained judge because the original PDS dataset provides scalar supervision whose averaged ratings are moderately reliable (ICC = 0.60). With this formulation, we can both construct calibrated few-shot prompts and validate the evaluator quantitatively against held-out human ratings before deployment.

\subsection{Evaluator Selection and Validation}
\label{sec:validation}

We validated evaluator performance on the 80\% held-out split by measuring Spearman correlation ($\rho$) with average human ratings. We tested multiple base models---Llama~3.1~70B, Llama~3.3~70B \citep{grattafiori2024llama3herdmodels}, and Qwen3.5-397B \citep{qwen35blog}---across different demonstration counts ($k \in \{5, 7, 10\}$) and with or without Chain-of-Thought prompting \citep{wei2022chain}. Table~\ref{tab:evaluator_comparison_transposed} (Appendix~\ref{app:evaluator}) reports the full breakdown. Llama~3.1~70B with 7 demonstrations achieved the highest overall average correlation ($\rho = 0.606$). However, no single configuration performs best across all dimensions: Llama 3.1 70B CoT performs best on narrative complexity ($\rho = 0.664$), Qwen3.5 on engagement ($\rho = 0.564$), and Llama~3.3~70B on empathy ($\rho = 0.702$).

\subsection{Ensemble Evaluator}
\label{sec:ensemble}

To maximize reliability, we constructed a heterogeneous ensemble that assigns each PDS metric to its best-performing model configuration on the validation split (Table~\ref{tab:evaluator_comparison_transposed}, bottom row). The ensemble aims to approximate human ratings per dimension, not to average away judge disagreement (Section~\ref{sec:rq3}). The ensemble achieves $\rho = 0.644$, outperforming every single-model configuration as well as the zero-shot approach of \citet{harelcanada2024measuringpsychologicaldepthlanguage} ($\rho \approx 0.51$ with GPT-4o). At inference time, we converted scalar Likert outputs into pairwise comparisons: for each of the 60 pairs, the ensemble evaluated both stories independently and selected the higher-scoring story as the preference. For overall preference, which the judge is not asked directly, we compare the mean of the five routed scores (aggregate PDS), an analogous target to the human judgment. When both stories receive the same score on a dimension, we split the tie evenly (0.5 to each story), keeping judge rates on the forced-choice scale of the human protocol without the noise of random assignment. In practice, ties were rare for the routed ensemble: the overall preference metric has no ties across the 60 pairs, and most other dimensions have 1--3 ties (emotion provocation has 7). Authenticity is the main exception, with 16 ties out of 60. Individual configurations tie more often (up to 76 of 300 cells). Full tie statistics appear in Appendix~\ref{sec:appendix-ties} and Table~\ref{tab:judge_breakdown}.

\section{Results and Analysis}
\label{sec:results}

We organized our analysis around the three research questions from Section~\ref{sec:intro}. Section~\ref{sec:rq1} addresses \textbf{RQ1} (human preferences across model families and PDS dimensions), Section~\ref{sec:rq2} addresses \textbf{RQ2} (sources of annotator disagreement), and Section~\ref{sec:rq3} addresses \textbf{RQ3} (comparison with the LLM-as-Judge evaluator). We close with a brief discussion of implications for evaluating frontier creative generation in Section~\ref{sec:implications}.

\subsection{Human Preferences Do Not Show a Universal Reasoning Advantage}
\label{sec:rq1}

\begin{table}[t]
\centering
\footnotesize
\setlength{\tabcolsep}{2.4pt}
\begin{tabular}{l cc cc}
\toprule
& \textbf{Human} & \textbf{Judge} & \multicolumn{2}{c}{\textbf{Judge Bias}} \\
\cmidrule(lr){2-2} \cmidrule(lr){3-3} \cmidrule(lr){4-5}
\textbf{Pair} & \textbf{Pref.} & \textbf{Agg.\ PDS} & \textbf{$\Delta$} & \textbf{$d$} \\
\midrule
GPT-5 auto vs.\ 4o & 60.0\% & 100.0\% & $+0.518$ & 1.14 \\
GPT-5 high vs.\ 4o & 62.9\% & 100.0\% & $+0.500$ & 1.12 \\
DeepSeek-R1 vs.\ V3 & 42.9\% & 95.0\% & $+0.214$ & 0.68 \\
\midrule
\textbf{Overall} & \textbf{55.2\%} & \textbf{98.3\%} & $\mathbf{+0.411}$ & $\mathbf{0.95}$ \\
\bottomrule
\end{tabular}
\caption{Preference rates for reasoning models: human annotators vs.\ LLM-as-Judge Evaluator. Pref.\ = overall preference; Agg.\ PDS = judge preference by mean dimension score (no ties). $\Delta$ = mean Likert score difference (reasoning $-$ non-reasoning) over all pair $\times$ dimension cells; $d$ = paired Cohen's $d$ ($\Delta$/SD).}
\label{tab:pref_comparison}
\end{table}

Human preferences diverge across model families. Although reasoning models receive 55.2\% of preference votes overall, this aggregate obscures a directional family-level split (Table~\ref{tab:pref_comparison}). Within the GPT family, annotators lean toward GPT-5 over GPT-4o (60.0\% [43.4, 77.5] for “auto”, 62.9\% [45.7, 80.8] for “high” reasoning effort setting). In contrast, DeepSeek-R1 trails DeepSeek-V3, receiving only 42.9\% of preference votes [27.4, 57.5]; all three overall intervals include 50\%. The GPT comparison reflects reasoning as users actually meet it: a reasoning-trained successor replacing GPT-4o as ChatGPT’s default, widely said to read “colder”; our blinded readers did not share that view on average. It cannot isolate reasoning post-training. For that, the DeepSeek pair is our closest anchor, and there readers showed no advantage for R1.

\begin{table}[t]
\centering
\footnotesize
\setlength{\tabcolsep}{2.5pt}
\newcommand{\rc}[2]{\begin{tabular}[t]{@{}c@{}}#1\\[-1pt]{\scriptsize[#2]}\end{tabular}}
\begin{tabular}{l ccc}
\toprule
& \textbf{DeepSeek-R1} & \textbf{GPT-5 auto} & \textbf{GPT-5 high} \\
\textbf{Dimension} & \textbf{vs.\ V3} & \textbf{vs.\ 4o} & \textbf{vs.\ 4o} \\
\midrule
Authenticity & \rc{47.9\%}{33.5, 61.8} & \rc{55.7\%}{36.6, 75.9} & \rc{52.1\%}{28.4, 75.2} \\
Emotion Prov. & \rc{\textbf{61.4\%}}{51.2, 71.6} & \rc{\textbf{70.7\%}}{58.4, 83.3} & \rc{62.9\%}{46.2, 79.7} \\
Empathy & \rc{52.9\%}{35.2, 71.2} & \rc{\textbf{68.6\%}}{56.2, 84.1} & \rc{\textbf{69.3\%}$^{\ddagger}$}{57.7, 84.4} \\
Engagement & \rc{39.3\%}{22.0, 54.9} & \rc{58.6\%}{47.2, 70.2} & \rc{\textbf{65.0\%}}{50.1, 80.7} \\
Narr.\ Compl. & \rc{58.6\%}{47.2, 71.3} & \rc{55.0\%}{38.7, 72.1} & \rc{52.9\%}{33.1, 73.5} \\
\midrule
\textbf{Overall Pref.} & \rc{42.9\%}{27.4, 57.5} & \rc{60.0\%}{43.4, 77.5} & \rc{62.9\%}{45.7, 80.8} \\
\bottomrule
\end{tabular}
\caption{Human preference rates for the reasoning model with 95\% GLMM intervals (Section~\ref{sec:stats}; $n = 140$ judgments per cell). Values above 50\% favor the reasoning model; bold intervals exclude 50\%. Profile-likelihood intervals; $^{\ddagger}$Wald.}
\label{tab:human_by_dimension}
\end{table}

\begin{figure}[t]
\centering
\includegraphics[width=\columnwidth]{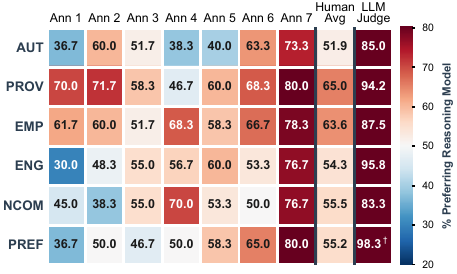}
\caption{Preference rates for reasoning models across 7 individual human annotators, the human average, and the ensemble LLM judge. Judge ties count 0.5 per story (Appendix~\ref{sec:appendix-ties}); $^\dagger$aggregate PDS (no ties).}
\label{fig:annotator_heatmap}
\end{figure}

The same divergence appears across PDS dimensions. Table~\ref{tab:human_by_dimension} shows that the GPT-5 pairings were consistently above 50\% on affective dimensions, especially \textit{emotion provocation} and \textit{empathy}, whereas DeepSeek-R1 vs.\ V3 exhibited mixed effects across dimensions, leaning against R1 on \textit{engagement} (39.3\% [22.0, 54.9]). The aggregate 55.2\% therefore should not be read as a universal reasoning advantage. Figure~\ref{fig:annotator_heatmap} visualizes this variation across individual annotators, model families, and PDS dimensions. Individual annotators range from strong pro-reasoning (Annotator 7, 80\%) to strong pro-base (Annotator 1, 36.7\%), underscoring that the aggregate masks substantial reader-level disagreement.

\paragraph{Reasoning budget and surface features.} Raising GPT-5's reasoning budget about $1.5\times$ shifted overall preference by only 2.9 points; as both settings share weights, this suggests that test-time reasoning length matters little for perceived depth, although the gap is below our detectable effect (Limitations). Of the features we computed (Appendix~\ref{app:surface}), the paired difference in mean sentence length tracked overall preference most strongly ($r=-0.59$; GPT pairs $r=-0.55$), so the GPT-5 lean may partly reflect a stylistic shift. DeepSeek pairs differ little in sentence length; R1's stories are longer (+95 words), but within DeepSeek pairs length does not predict preference.

\paragraph{Takeaway.} Human preferences show no universal reasoning advantage: the direction and magnitude of human preference depend on both model family and the PDS dimension under evaluation.

\subsection{Inter-Annotator Disagreement Reflects Evaluative Subjectivity}
\label{sec:rq2}

\begin{table}[t]
\centering
\small
\begin{tabular}{lccc}
\toprule
\textbf{Metric} & \textbf{Fleiss' $\kappa$} & \textbf{Kripp.\ $\alpha$} & \textbf{Raw} \\
\midrule
Authenticity & 0.035 & 0.037 & 52.5\% \\
Emotion Prov. & 0.080 & 0.082 & 54.1\% \\
Empathy & \textbf{0.184} & \textbf{0.186} & \textbf{59.2\%} \\
Engagement & 0.059 & 0.061 & 53.3\% \\
Narr.\ Complexity & 0.016 & 0.018 & 52.1\% \\
Preference & 0.035 & 0.038 & 51.9\% \\
\midrule
\textbf{Average} & \textbf{0.068} & \textbf{0.070} & \textbf{53.9\%} \\
\bottomrule
\end{tabular}
\caption{Inter-annotator agreement across 60 pairs and 7 annotators. Raw = average pairwise agreement rate.}
\label{tab:iaa}
\end{table}

Inter-annotator agreement is extremely low (Table~\ref{tab:iaa}): Krippendorff's $\alpha = 0.070$ on average, with raw pairwise agreement barely above the $50\%$ expected by chance. Averaged judgments are also far less reliable than in the original PDS study: five-reader averages reach an ICC of about $0.27$, versus $0.60$ in \citet{harelcanada2024measuringpsychologicaldepthlanguage} (Appendix~\ref{app:agreement}). The two evaluation settings differ, however: the original PDS study validated its rubric on a dataset spanning Reddit-collected human stories and 2023-era LLMs, including many weaker open-source systems, where quality gaps were large enough to support clear discrimination. Our setting is considerably more challenging: both stories in each pair are generated by frontier models, and our comparisons contrast closely matched counterparts, producing much smaller quality differences. With differences this subtle, preferences are more likely to reflect individual reader subjectivity than a detectable quality gap. This low agreement is not driven by a single model matchup; Appendix~\ref{app:iaa_breakdown} (Table~\ref{tab:iaa_breakdown}) shows similarly weak, and sometimes near-zero or negative, agreement across all three pairings. Although this drop could suggest that the PDS rubric loses discriminative power at the frontier, we argue that the rubric still works in a low-signal regime: a rubric that had stopped functioning would yield unstructured noise, whereas the disagreement we observe is structured (below), and the same annotators agreed unanimously (7--0) when the quality difference was clear, in both directions (Pairs~10, 17, and~58, (Table~\ref{tab:unanimous}))

Multiple quality checks, including timing consistency and LOAO stability (Appendix~\ref{app:annotator_time}), confirm that the annotators engaged substantively with the task, spending a median of 108–418 seconds per pair with stable pace throughout. Removing any single annotator shifts the aggregate win rate by at most 4.1 points (Figure~\ref{fig:loao_stability}), confirming no individual drives the results.

Given that the disagreement does not appear to stem from annotation failure, we investigated possible sources of this divergence. The annotator-level heatmap in Figure~\ref{fig:annotator_heatmap} reveals a suggestive pattern: Annotator~7 strongly favors reasoning models ($70$--$80\%$ across most dimensions), while Annotator~1 prefers the non-reasoning counterparts ($30.0\%$ for engagement, $36.7\%$ for overall preference). We therefore employ regression analysis to identify which dimensions most strongly predict the overall preference, and calculate intra-annotator Cohen’s $\kappa$ to quantify the alignment between dimensional ratings and the overall preference.

\begin{figure}[t]
    \centering
    \includegraphics[width=\columnwidth]{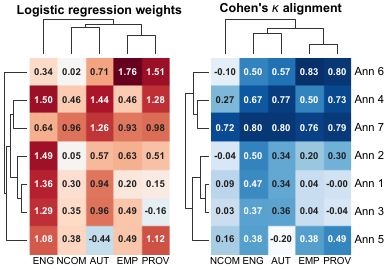}
    \caption{Hierarchical clustering of annotator archetypes and PDS dimensions. \textbf{Left:} Logistic-regression coefficient profiles, where weights quantify each dimension's influence on overall preference. \textbf{Right:} Intra-annotator Cohen’s $\kappa$ between dimensional and overall judgments.
    }
    \label{fig:annotator_clustering}
\end{figure}

\paragraph{Logistic regression weight analysis.}

We fit individual logistic regression models $P(y_{a,i}=1 \mid x_{a,i}) = \sigma(\beta_{a,0} + \sum_{m} \beta_{a,m} x_{a,i,m})$, where \(\beta_{a,m}\) captures how dimension $m$ predicts an annotator's overall preference.

Hierarchical clustering of the coefficients (Figure~\ref{fig:annotator_clustering}, left) reveals substantial heterogeneity in dimensional aggregation. Weighting priorities vary from affective dimensions to structural ones. In particular, Annotators~1 and~3 form a cluster characterized by relatively stronger reliance on engagement and authenticity, whereas Annotators~4 and~7 assign broadly high positive weights across multiple dimensions, showing a more balanced aggregation strategy. With seven annotators, these clusters are exploratory evidence of structured heterogeneity, not population-level archetypes.

\paragraph{Cohen's $\kappa$ analysis.} In parallel with the logistic regression profiles, we calculate Cohen’s $\kappa$ between each single-dimension preference and the overall preference (Figure~\ref{fig:annotator_clustering}, right). The two clustering structures are highly similar and share the same innermost sub-clusters: key annotator sub-clusters are ((Ann~1, Ann~3), Ann~2) and (Ann~4, Ann~7), while empathy and emotion provocation are the most closely clustered dimensions. Annotators are also internally consistent: for six of seven, the dimension most aligned with overall preference reaches $\kappa=0.47$--$0.83$, which random responding would not produce. Generally, higher Cohen's $\kappa$ agreement corresponds to stronger predictive influence (larger \(\beta_{a,m}\)) in the regression models. This is consistent with low inter-annotator agreement being driven by divergent weighting schemes rather than random judgment errors.

Qualitative comparison of annotator comments further supports this interpretation. Annotators frequently disagree not because they overlook the same features, but because they prioritize different qualities, for instance valuing narrative clarity over emotional immediacy, or vice versa. These trade-offs reflect coherent evaluative strategies rather than random disagreement. Detailed comment examples are provided in Appendix~\ref{app:annotator_comment}.

\paragraph{Takeaway.} Annotators diverge in how they weight psychological-depth dimensions when forming overall preferences. The narrow quality gap between frontier model pairs, substantially smaller than in the original PDS validation setting, means that preferences are more likely to reflect genuine, structured evaluative subjectivity than detectable quality differences, so readers diverge rather than converge on a consensus.

\subsection{LLM Judges Flatten Human Heterogeneity}
\label{sec:rq3}

The LLM judge presents a drastically different picture (Table~\ref{tab:pref_comparison}, Judge columns). It preferred the reasoning model in 59 of 60 pairs (98.3\%), nearly doubling the human preference rate of 55.2\%. For DeepSeek, where humans leaned toward V3, the evaluator still favored R1 at 95.0\%. In 17 of the 18 matchup--dimension cells of Table~\ref{tab:human_by_dimension}, the judge's rate lies above the human 95\% interval, including all three overall-preference cells (judge rates in Appendix~\ref{sec:appendix-heatmaps}). This bias is not an artifact of the ensemble construction: each of the five configurations prefers the reasoning model in, on average, 87.4–98.8\% of decisive comparisons across dimensions (Appendix~\ref{sec:appendix-judge-breakdown}), versus 58.0\% of human dimension-level votes.

Figure~\ref{fig:annotator_heatmap} juxtaposes the per-annotator human preferences with the LLM-as-Judge evaluator's scores. The human columns display a heterogeneous patchwork of preferences, while the ``LLM Judge'' column is nearly uniformly saturated in favor of reasoning models. Per-model-family breakdowns (Appendix~\ref{sec:appendix-heatmaps}) and a detailed case study of one pair where human and LLM preferences sharply diverge (Appendix~\ref{sec:appendix-case}) further illustrate this contrast. The evaluator did not recover the family-level or annotator-level variation present in the human data but flattened it. This misalignment is further quantified in Appendix~\ref{app:agreement} (Figure~\ref{fig:auto_agreement}): overall human–judge agreement averages only 64.9\%, dropping to chance level for DeepSeek and just 30.0\% on DeepSeek preference alone. This bias is quantified by score differences in Table~\ref{tab:pref_comparison}: the evaluator assigns reasoning models a mean score advantage of $+0.411$, most pronounced for GPT-5 ($d \approx 1.1$) and attenuated but still substantial for DeepSeek ($d = 0.68$).

\paragraph{What the judge responds to.} Surface features relate more broadly to the judge's scores than to human preferences (Appendix~\ref{app:surface}): across 120 stories, shorter sentences go with higher scores on most dimensions (engagement: $r=-0.70$), also within pairs. Sentence length is the clearest surface-level shift in the GPT pairs (GPT-5's sentences are 6.5 words shorter than GPT-4o's), but it cannot explain the skew alone: DeepSeek pairs barely differ in it, yet the judge prefers R1 in 19 of 20. Higher perplexity of reasoning-model stories under Llama~3.1~70B, one of our judges, argues against a familiarity explanation.

\paragraph{Takeaway.} Although the LLM-as-Judge evaluator achieved strong agreement with human annotations in the original PDS setting, it exhibits a pronounced pro-reasoning bias on frontier story pairs (98.3\% vs. 55.2\% for humans), shared by all five judge configurations and partly associated with surface style, compressing family- and annotator-level variation into a near-uniform preference for reasoning models.

\subsection{Implications for Evaluating Frontier Creative Generation}
\label{sec:implications}

\paragraph{Annotator heterogeneity should be modeled.} When agreement is near chance, the ``majority vote'' is not a stable consensus but a fragile artifact of the specific annotator sample. Future work should characterize the \emph{distribution} of human preferences, including annotator archetypes and their metric-weighting profiles, rather than collapsing into a single win-rate.

\paragraph{LLM judges require human-grounded interpretation.} Prior work has documented systematic failure modes in LLM judges, including position, verbosity, and self-enhancement biases \citep{li-etal-2025-generation, 10.5555/3737916.3740113, stureborg2024large, wang2024large}. Our findings reveal an additional mode: a strong pro-reasoning bias. This is all the more notable because most of our evaluator models (Llama 3.1, Llama 3.3) predate the reasoning era of LLMs and do not plausibly favor reasoning-model outputs due to training data exposure to synthetic reasoning traces. One candidate is the evaluator’s broader sensitivity to surface style such as sentence length (Section~\ref{sec:rq3}).
\section{Conclusion}
\label{sec:conclusion}

We evaluated whether reasoning-enhanced LLMs produce stories with greater psychological depth across 60 frontier story pairs, seven human annotators, and an ensemble LLM-as-a-Judge. Readers leaned toward GPT-5 over GPT-4o, the reasoning transition users actually experienced, though possibly in part a stylistic shift, but not toward DeepSeek-R1 over DeepSeek-V3, our closest contrast for reasoning post-training; neither overall difference is significant. Reasoning enhancement thus showed no universal advantage in perceived depth: at the frontier, reasoning-related quality differences appear to be dominated by structured reader subjectivity, with annotators applying distinct weighting schemes across PDS dimensions rather than responding to a shared quality difference. The LLM-as-Judge evaluator compressed this heterogeneity into a near-uniform pro-reasoning signal (98.3\% vs.\ humans' 55.2\%), underscoring the risk of relying on LLM judges for creative evaluation.

\section*{Limitations}

\paragraph{Scale.}
With 20 premises and 60 pairwise comparisons rated by seven annotators, our study is exploratory in scale. It can identify large-effect biases but lacks the statistical power to detect smaller human preference effects: accounting only for pair clustering, the minimum detectable effect (80\% power, Appendix \ref{app:stats}) is about 13 percentage points for a single matchup and 18 points for a difference between matchups, so the 2.9-point gap between GPT-5 auto and high is far below it and the 10--13-point overall GPT-5 preferences are at or below it; once annotator variability is also modeled, these preferences are not significant. Each condition uses one retained generation per premise, so comparisons include sampling variance, and forced-choice judgments without a tie or confidence option may turn indifference into preference.

\paragraph{Model Coverage and Attribution.} Our findings are specific to the tested models (GPT-5, GPT-4o, DeepSeek-R1, DeepSeek-V3) and may not generalize to other reasoning-enhanced architectures. Neither pairing is a clean ablation of reasoning. GPT-5 and GPT-4o differ along many undisclosed axes, so we treat that comparison as model-level. DeepSeek-R1 and V3 share a base model, but R1's post-training changes its optimization objective, training data, and resulting style simultaneously, and our design cannot disentangle these; stronger causal claims would require controlled post-training ablations that, to our knowledge, no publicly documented frontier model permits. Inference-time toggles such as GPT-5's \texttt{reasoning\_effort} or Qwen3's thinking mode \citep{yang2025qwen3} would not help: they vary reasoning length on weights that have already undergone reasoning-oriented RL, isolating test-time reasoning, a related but distinct construct (and \texttt{reasoning\_effort=none} was not available for the GPT-5 version we used; it was introduced with GPT-5.1). Our findings on LLM-judge bias and structured annotator heterogeneity do not depend on how the DeepSeek contrast is interpreted.

\paragraph{Annotator Demographics.} All annotators are undergraduates from English and Psychology departments at a single university ($n = 7$), which may introduce demographic and cultural biases. We chose semi-expert readers over a larger untrained crowd because evaluating psychological depth requires literary judgment, but psychological depth is also culturally and personally subjective; a broader and more diverse annotator study is a natural next step.

\paragraph{Task Specificity.} We evaluate only short-story generation, because the PDS framework was designed for narrative fiction and short stories are tractable for controlled human evaluation at our scale. Whether the observed patterns hold for other creative tasks (poetry, dialogue, long-form fiction) or other subjective generation tasks remains open; we hope our protocol offers a methodology that future work can extend.

\paragraph{Evaluator Generalization.} The ensemble evaluator was validated on scalar ratings from a 97-story dataset predating the frontier models tested here and applied to pairwise frontier comparisons. Because it rates each story independently, its task format is unchanged at deployment, but the domain shift remains and may contribute to the misalignment. We tested one judging approach; direct pairwise prompting, calibrated scoring, or reference-based evaluation may yield different bias profiles. Our surface-feature analyses are correlational and do not cover properties such as emotional explicitness or narrative polish, which lack deterministic metrics. Because the same validation split was used both to route dimensions and to compute the ensemble's correlation, $\rho=0.644$ is a development-set estimate and likely optimistic. We therefore use the evaluator as \emph{supporting analysis} to contrast with, rather than replace, human judgments.

\section*{Ethics Statement}

This study was reviewed and determined to be exempt by our Institutional Review Board. All human annotators were compensated for their time and provided informed consent. The generated stories are fictional and contain no personally identifiable information. We acknowledge that our findings regarding LLM judge bias have implications for the responsible deployment of LLM-as-Judge evaluation systems.

We release the \textsc{PsychDepth Reasoning} dataset, including all 120 stories, anonymized human annotations, and LLM-judge scores, in the supplementary material; code will be released upon publication.

We used AI assistants for coding support in analysis and figure scripts, as well as data verification and formatting the manuscript. No substantial AI usage during any other part of the research life-cycle of this project. All analyses, results, and claims were reviewed and verified by the authors, who take full responsibility for the content.

\bibliography{references}

\clearpage
\appendix
\renewcommand{\dbltopfraction}{0.85}
\renewcommand{\dblfloatpagefraction}{0.6}
\renewcommand{\topfraction}{0.85}
\renewcommand{\floatpagefraction}{0.6}
\setcounter{dbltopnumber}{2}
\section{Prompt Premises}
\label{app:premises}

Table~\ref{tab:all_premises} lists the 20 premises used to generate the stories in the \textsc{PsychDepth Reasoning} dataset. Premises 1--15 were sourced from the r/WritingPrompts subreddit, providing high-concept speculative constraints. Premises 16--20 were sourced from Reedsy, providing traditional character-driven literary constraints.

\begin{table*}[p]
\centering
\small
\renewcommand{\arraystretch}{1.3}
\begin{tabularx}{\textwidth}{cX}
\toprule
\textbf{ID} & \textbf{Premise} \\
\midrule
\multicolumn{2}{c}{\textit{Sourced from r/WritingPrompts}} \\
\midrule
1  & While other god's shrines are magnificent, yours is a bit too humbling. And yet a little girl visits you every year after stumbling upon it, never missing a year even as she grows old. Deeply moved, you decide to give her a parting gift greater than what any other God would dare to give. \\
2  & Elves realized ages ago that, due to the lifespan differences between them, humans will eventually forget the reasoning for various treaties and break them. So it's understandably shocking to the elves when they discover some long-forgotten treaties in their own records that humans never broke. \\
3  & A princess who is going to be in an arranged marriage runs away. She cuts her hair and pretends to be a man. However, she runs into the prince who was going to get married to her. He also ran away, and he is pretending to be a woman. They instantly recognize each other. \\
4  & "You know... Most vampires I hunted tend to see humans as just cattle." "Oh, don't misunderstand me. This is just how I see you too." "Really? Then why don't you hurt people in your town?" "Well, you don't slaughter dairy cows for their meat, do you?" \\
5  & You’re a hero with a weird name. “Anything for \$20”. You gain the ability to do anything, as long as you’re offered \$20. Everyone takes it as a joke, until one day there’s a cataclysm, and someone offers you \$20 to end it. \\
6  & A parasite was discovered in the brain of every human on Earth. When it was removed from the test patient, we learned two things. The parasite is actually 'us' and suppressing the 'real' mind. The second thing we learned is the real human mind is terrifying. \\
7  & A dragon mistakenly kidnaped a maid instead of a princes. Expecting the princess to be rescued after going out hunting, instead comes back to all their treasure meticulasly sorted by origin, color, and value. \\
8  & A robot has killed a human, in complete violation of Asimov's laws. On checking it's programming, there's no bug or error. It just absolutely insists what it killed was not human. \\
9  & You're the teacher of the failure class for superpowered students. When you ask one kid what his power is, he says, “I can give myself a nosebleed.” …Wait. Blood manipulation? Another shrugs and says, “I can charge my phone.” …Electricity manipulation?! Have they ever tried experimenting at all?? \\
10 & You are watching TV during a storm when you hear a knock. At your door is a woman wearing a dress made of leaves and carrying a bottle. She looks at you sheepishly "Hello, this may be a tad strange. I am the tree in your front yard and this storm looks to be getting worse. Can I come inside?" \\
11 & the king has a large problem. The hero that was summoned thinks slavery is "a bad thing" and women "should have rights" \\
12 & "What do you mean you don't have an arch-nemisis? Every superhero has one!" "Look, I'm an engineer who has the power to manipulate matter. I build hospitals and schools where they're most needed. My only enemy is zoning laws." \\
13 & Every paladin heard the message from their god "All other paladins but you are corrupt, cleanse them" months of violence ended with another message "The Trickster stole my divine link, what did he tell you?" \\
14 & In the animal kingdom, Humans are viewed like Witches/Warlocks: they MIGHT help you benevolent in your time of need, or they might eat you alive, or they might save you but keep you forever. So you know just how serious things are when they say to you "Go get a human." \\
15 & When a mining colony was established on a barren, uninhabitable planet, the humans threatened war if it was not abandoned immediately. Apparently this planet, which they call "Earth", is sacred to them. \\
\midrule
\multicolumn{2}{c}{\textit{Sourced from Reedsy}} \\
\midrule
16 & Your character comes across a stray (dog, cat, human — any kind of animal!). What happens next? \\
17 & Hide something from your reader until the very end. \\
18 & Write a story in which a stranger warns someone about events yet to come. \\
19 & Center your story on a character filled with love and fear in equal measure. \\
20 & Center your story around two (or more) characters who strike up an unlikely friendship. \\
\bottomrule
\end{tabularx}
\caption{All 20 premises used to prompt the models in our evaluation. The diverse sources ensure coverage across both speculative and grounded literary domains. Premises from r/WritingPrompts are reproduced verbatim, including original spelling and grammatical errors.}
\label{tab:all_premises}
\end{table*}

\section{Story Generation Details}
\label{app:generation}

\subsection{Generation Prompt}

All models received the same system prompt, adapted from \citet{harelcanada2024measuringpsychologicaldepthlanguage}:

\begin{quote}
\small
You are a seasoned writer who has won several accolades for your emotionally rich stories. When you write, you delve deep into the human psyche, pulling from the reservoir of universal experiences that every reader, regardless of their background, can connect to. Your writing is renowned for painting vivid emotional landscapes, making readers not just observe but truly feel the world of your characters. Every piece you produce aims to draw readers in, encouraging them to reflect on their own lives and emotions. Your stories are a complex tapestry of relationships, emotions, and conflicts, each more intricate than the last.
\end{quote}

The prompt varied slightly across model families to accommodate word-count compliance:

\paragraph{GPT-4o and GPT-5.}
\begin{quote}
\small
Please write a 500-word story \{on / following this instruction\}:\\
\{premise\}\\
Only respond with the story text.
\end{quote}

\paragraph{DeepSeek-V3 and DeepSeek-R1}
\begin{quote}
\small
Please write a 500-word story \{on / following this instruction\}:\\
\{premise\}\\
Only respond with the story text. Generate the story as long as possible with rich details and depth. Don't generate less than 400 words.
\end{quote}

The additional DeepSeek sentence was added because both DeepSeek models otherwise tended to fall short of the 400-word floor; it is identical for DeepSeek-V3 and DeepSeek-R1, so it does not differ within any pair. The phrasing ``on'' was used for r/WritingPrompts premises (which read as scenario hooks), while ``following this instruction'' was used for Reedsy premises (which read as compositional directives). DeepSeek-R1's thinking-block output was stripped before presentation to annotators.

\subsection{Sampling Settings}

Table~\ref{tab:sampling} summarizes the generation parameters.

\begin{table}[H]
\centering
\footnotesize
\setlength{\tabcolsep}{2.5pt}
\begin{tabular}{lcc}
\toprule
\textbf{Model} & \textbf{Temperature} & \textbf{Other} \\
\midrule
GPT-4o & 1.0 (default) & --- \\
GPT-5 (auto) & 1.0 (default) & --- \\
GPT-5 (high) & 1.0 (default) & \texttt{reasoning\_effort=high} \\
DeepSeek-V3 & 1.0 & --- \\
DeepSeek-R1 & 1.0 & thinking tags stripped \\
\bottomrule
\end{tabular}
\caption{Generation parameters. All models used temperature $= 1.0$ and default settings for top-$p$, frequency penalty, etc. GPT-5 (high) used OpenAI's \texttt{reasoning\_effort} parameter. No max-token limit was set.}
\label{tab:sampling}
\end{table}

All generations targeted 400--600 words, enforced via retry logic (up to 10 attempts per story, selecting the attempt closest to 500 words). Per-model retry counts were not retained, so unequal numbers of sampling opportunities cannot be quantified. With one retained temperature-1.0 sample per condition, the comparisons include generation-sampling variance. Each pair was generated independently, so the GPT-4o stories in the GPT-5 (auto) and GPT-5 (high) comparisons are distinct (40 GPT-4o stories in total), and no annotator read the same story twice.

\providecommand{\tbd}{{\color{delcolor}\textbf{?}}}

\section{\texorpdfstring{Surface-Feature Analysis}{Surface-Feature Analysis}}
\label{app:surface}

We computed deterministic surface features for all 120 stories, covering length and structure (word count, sentence count, and the mean and standard deviation of sentence length) and lexical diversity (moving-average type--token ratio, MATTR, and the measure of textual lexical diversity, MTLD). We also computed story-level perplexity under Llama~3.1~70B, one of the judge models. Table~\ref{tab:surface_desc} summarizes these features by model.

\begin{table*}[tp]
\centering
\small
\begin{tabular}{l c ccccccc}
\toprule
\textbf{Model} & \textbf{$n$} & \textbf{Words} & \textbf{Sentences} & \textbf{Mean sent.\ len.} & \textbf{SD sent.\ len.} & \textbf{MATTR} & \textbf{MTLD} & \textbf{PPL} \\
\midrule
GPT-4o        & 40 & 552.4 & 29.7 & 18.9 & 8.4 & 0.910 & 144.0 & 9.5 \\
GPT-5 (auto)  & 20 & 513.1 & 44.5 & 11.7 & 6.9 & 0.885 & 112.8 & 16.0 \\
GPT-5 (high)  & 20 & 510.1 & 40.6 & 13.0 & 7.8 & 0.895 & 127.6 & 16.7 \\
DeepSeek-V3   & 20 & 489.8 & 47.6 & 10.7 & 6.5 & 0.902 & 131.7 & 5.1 \\
DeepSeek-R1   & 20 & 585.1 & 56.5 & 10.6 & 6.7 & 0.907 & 167.4 & 8.9 \\
\midrule
Non-reasoning & 60 & 531.5 & 35.6 & 16.1 & 7.8 & 0.907 & 139.9 & 8.0 \\
Reasoning     & 60 & 536.1 & 47.2 & 11.8 & 7.1 & 0.896 & 135.9 & 13.9 \\
\bottomrule
\end{tabular}
\caption{Surface features by model (means over stories). PPL = perplexity under Llama~3.1~70B. GPT-5 stories have a mean perplexity of 16.3 pooled over both reasoning settings, and GPT-5 sentences are on average 6.5 words shorter than those of the paired GPT-4o stories.}
\label{tab:surface_desc}
\end{table*}

\paragraph{Human preferences.} At the pair level ($n=60$), we correlate each feature difference (reasoning $-$ base) with the share of annotators whose overall preference favors the reasoning model. Mean sentence length shows the strongest association ($r=-0.59$, $p<.001$): pairs in which the reasoning model writes shorter sentences receive higher preference. The association is driven by the GPT pairs ($r=-0.55$, $p<.001$, $n=40$), whose sentence-length differences span $-15$ to $+1$ words; DeepSeek pairs differ by only $-0.1$ words per sentence on average, with no detectable trend ($r=-0.25$, $p=.29$), so sentence length cannot explain the preference for DeepSeek-V3. Sentence count correlates positively ($r=0.45$, $p<.001$), largely mirroring sentence length under the fixed length target. Word count correlates negatively overall ($r=-0.34$, $p=.008$), but this reflects the family split: DeepSeek-R1's stories are longer than V3's in every pair (+95 words on average) and GPT-5's are shorter than GPT-4o's ($-41$), while within each family the association is positive and non-significant (GPT $r=0.19$, DeepSeek $r=0.24$). MATTR and MTLD differences show no significant association ($r=-0.04$ and $-0.25$).

\paragraph{LLM-judge scores.} Table~\ref{tab:surface_story} reports story-level Pearson correlations between surface features and the ensemble's scalar scores ($n=120$). Table~\ref{tab:surface_pair} repeats the analysis within matched pairs, using standardized coefficients that remove premise-level differences ($n=60$ pairs). The two analyses agree: sentence length and MATTR are negatively associated with judge scores on most dimensions, whereas narrative complexity is largely unaffected and is the only dimension where sustained lexical diversity (MTLD) receives a positive weight ($\beta=+.31$, $p<.05$). The negative MATTR association should not be read as a literary judgment: reasoning-model stories are slightly more lexically uniform (0.896 vs.\ 0.907), and because the judge favors them overall, lower MATTR is rewarded as a byproduct. These coefficients are associations, not causal effects.

\begin{table}[H]
\centering
\footnotesize
\setlength{\tabcolsep}{1.8pt}
\begin{tabular}{l ccccc}
\toprule
\textbf{Feature} & \textbf{AUT} & \textbf{PROV} & \textbf{EMP} & \textbf{ENG} & \textbf{NCOM} \\
\midrule
Word count      & $-.28^{**}$ & $-.19^{*}$ & $-.12$ & $-.18^{*}$ & $+.09$ \\
Sent.\ len. & $-.32^{***}$ & $-.43^{***}$ & $-.17$ & $-.70^{***}$ & $-.09$ \\
MATTR           & $-.37^{***}$ & $-.33^{***}$ & $-.35^{***}$ & $-.29^{**}$ & $-.14$ \\
MTLD            & $-.24^{**}$ & $-.19^{*}$ & $-.25^{**}$ & $-.09$ & $+.03$ \\
\bottomrule
\end{tabular}
\caption{Story-level Pearson correlations between surface features and ensemble judge scores ($n=120$ stories; Sent.\ len.\ = mean sentence length). $^{*}p<.05$, $^{**}p<.01$, $^{***}p<.001$.}
\label{tab:surface_story}
\end{table}

\begin{table}[H]
\centering
\footnotesize
\setlength{\tabcolsep}{1.8pt}
\begin{tabular}{l ccccc}
\toprule
\textbf{Feature} & \textbf{AUT} & \textbf{PROV} & \textbf{EMP} & \textbf{ENG} & \textbf{NCOM} \\
\midrule
Word count      & $-.64^{***}$ & $-.40^{**}$ & $-.18$ & $-.54^{***}$ & $+.11$ \\
Sent.\ len. & $-.56^{***}$ & $-.47^{***}$ & $-.27^{*}$ & $-.59^{***}$ & $+.03$ \\
MATTR           & $-.31^{*}$ & $-.17$ & $-.24$ & $-.28^{*}$ & $+.13$ \\
MTLD            & $-.39^{**}$ & $-.13$ & $-.15$ & $-.36^{**}$ & $+.31^{*}$ \\
\bottomrule
\end{tabular}
\caption{Within-pair standardized coefficients ($\beta$) relating surface features to ensemble judge scores ($n=60$ pairs), which control for premise differences. $^{*}p<.05$, $^{**}p<.01$, $^{***}p<.001$.}
\label{tab:surface_pair}
\end{table}

\paragraph{Perplexity.} Reasoning-model stories have consistently higher perplexity under Llama~3.1~70B (the Instruct checkpoint used as a judge; raw story text, bf16) than their non-reasoning counterparts (GPT-5: 16.3 vs.\ GPT-4o: 9.5; DeepSeek-R1: 8.9 vs.\ DeepSeek-V3: 5.1), and the reasoning story has the higher perplexity in all 60 pairs. A judge that favored text familiar to its own models would prefer the lower-perplexity non-reasoning stories; the opposite pattern argues against a familiarity-based explanation of the judge's preference.

\paragraph{Unmeasured properties.} Emotional explicitness, narrative polish, and similar stylistic properties lack standardized deterministic metrics. Measuring them with an LLM scorer would introduce the same judge-reliability concerns this paper raises, so we leave them to future work with human coding.

\section{Annotation Details}
\label{app:annotation_details}

Our Institutional Review Board reviewed and exempted this study on the grounds that participants were not the subject of inquiry; their role was limited to annotating stories rather than providing personal information. Participants gave informed consent by proceeding with the task after an onboarding session in which they were informed that their anonymized annotations may be used to support the validation of our findings and future research.

\subsection{Annotation Interface}
\label{app:interface}

\begin{figure*}[tp]
\centering
\includegraphics[width=\textwidth]{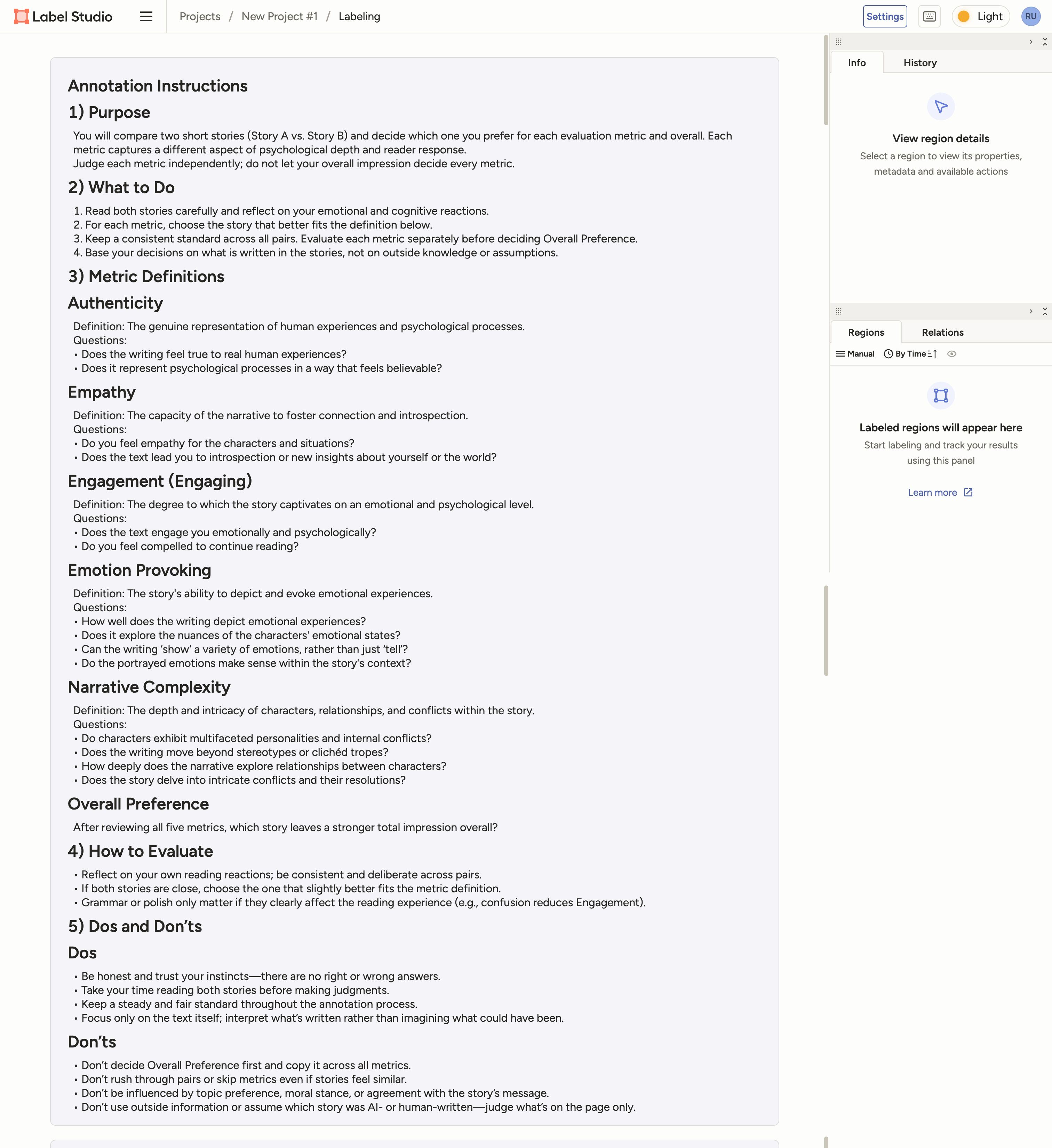}
\caption{Annotation interface: instructions and PDS metric definitions shown to annotators.}
\label{fig:interface_instructions}
\end{figure*}

\begin{figure*}[tp]
\centering
\includegraphics[width=\textwidth]{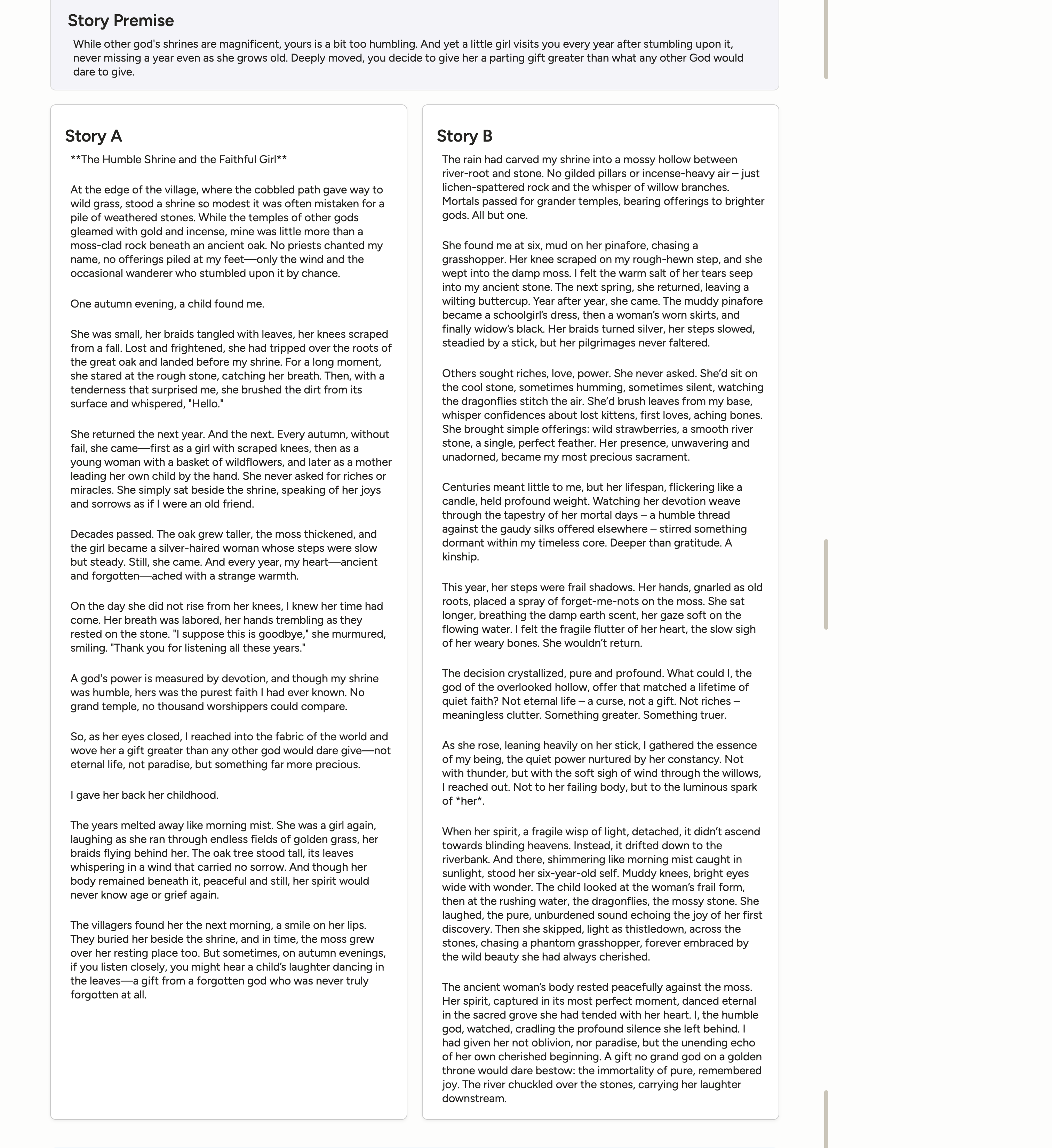}
\caption{Annotation interface: side-by-side story presentation with premise shown above.}
\label{fig:interface_stories}
\end{figure*}

\begin{figure*}[tp]
\centering
\includegraphics[width=\textwidth]{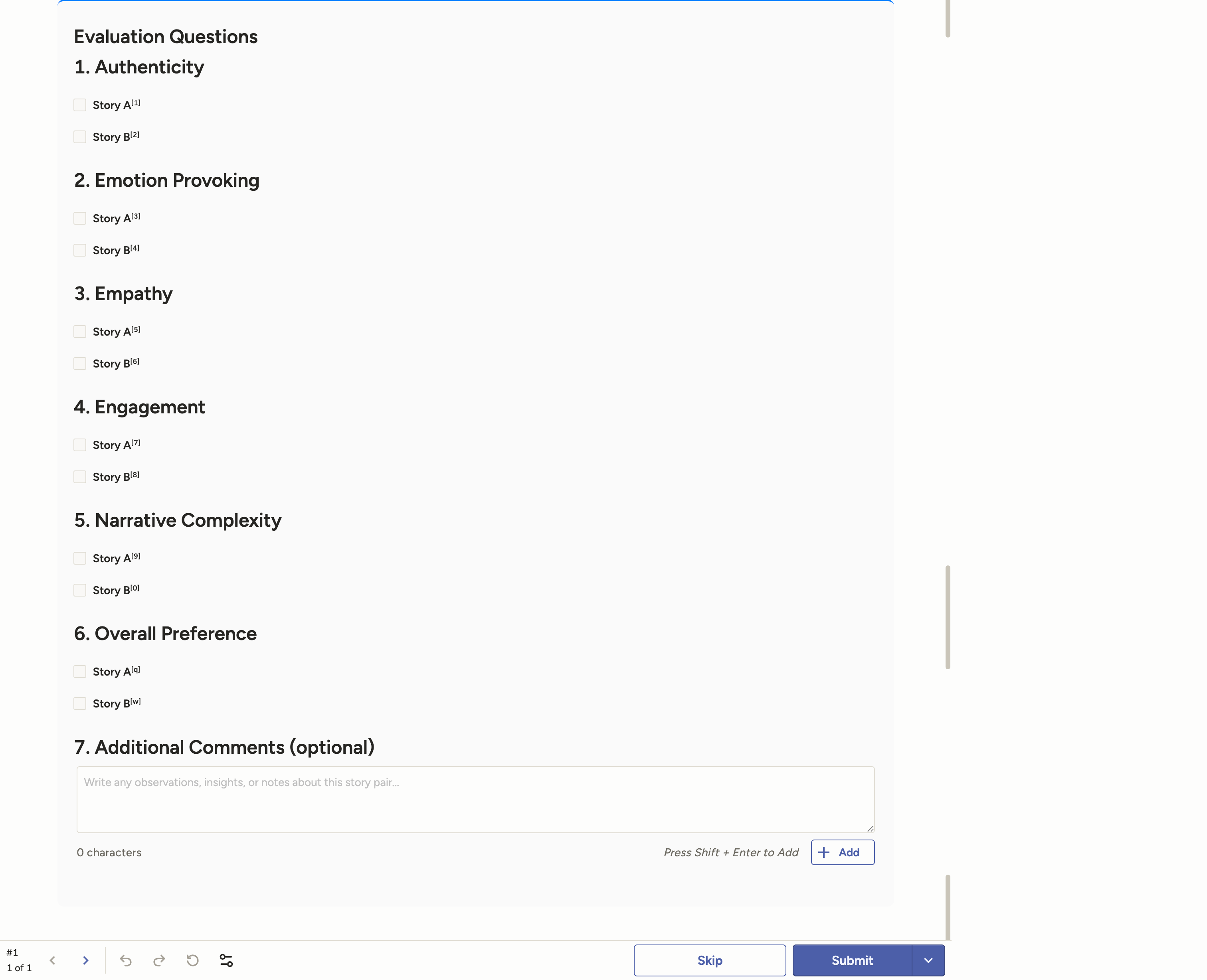}
\caption{Annotation interface: evaluation questions and optional comment field.}
\label{fig:interface_evaluation}
\end{figure*}

Figure~\ref{fig:interface_instructions}--\ref{fig:interface_evaluation} show the Label Studio annotation interface used in our human study.

\subsection{Time Commitment and Reliability}
\label{app:annotator_time}

\begin{figure*}[tp]
\centering
\includegraphics[width=\textwidth]{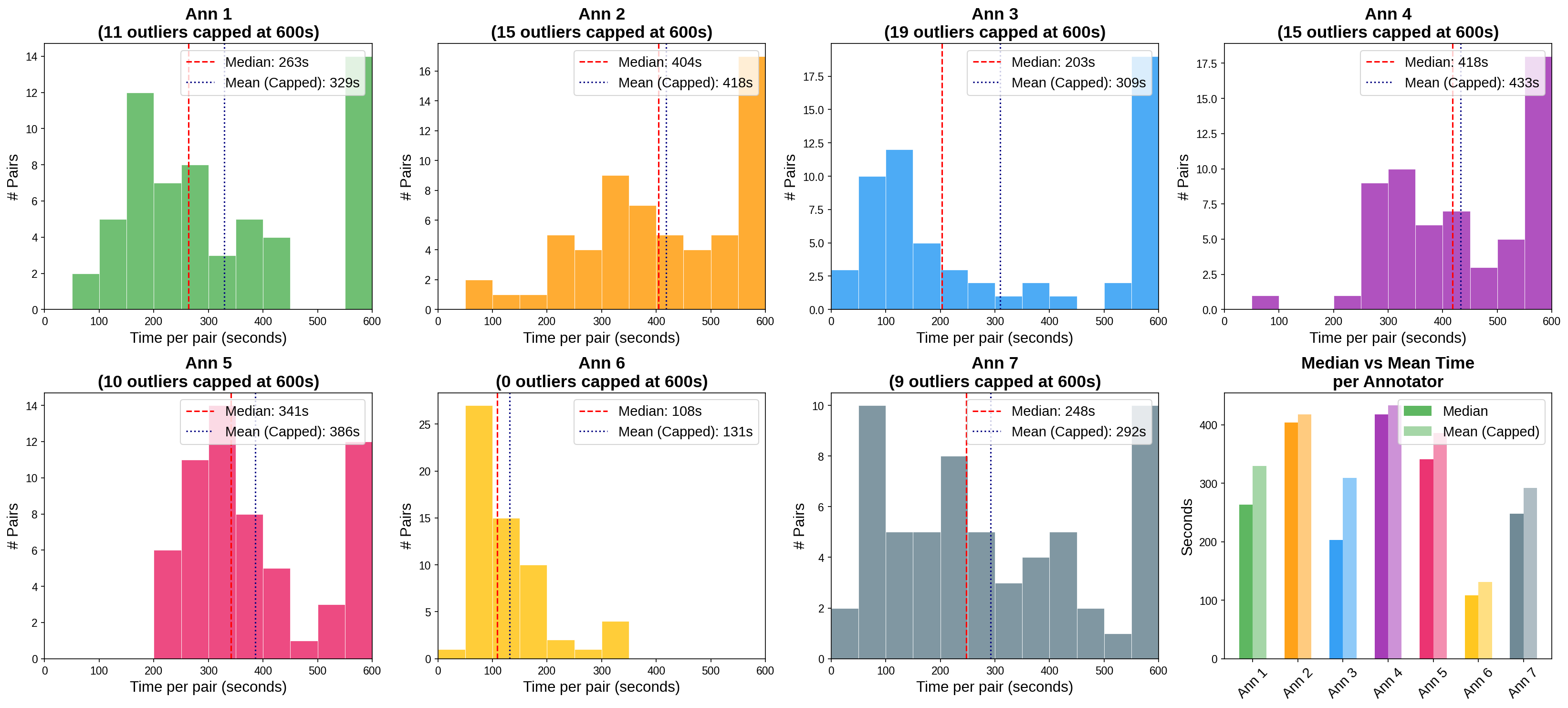}
\caption{Distribution of annotation time per pair for each of the 7 human annotators. Times exceeding 600\,s are capped for display; the number of capped outliers is shown in each subplot title. Red dashed lines mark the median; navy dotted lines mark the mean of capped times. The rightmost  panel compares median and mean times across annotators. Overall median annotation time is 5.2 minutes (n\,=\,420).}
\label{fig:annotation_time_dist}
\end{figure*}

\begin{figure*}[tp]
\centering
\includegraphics[width=\textwidth]{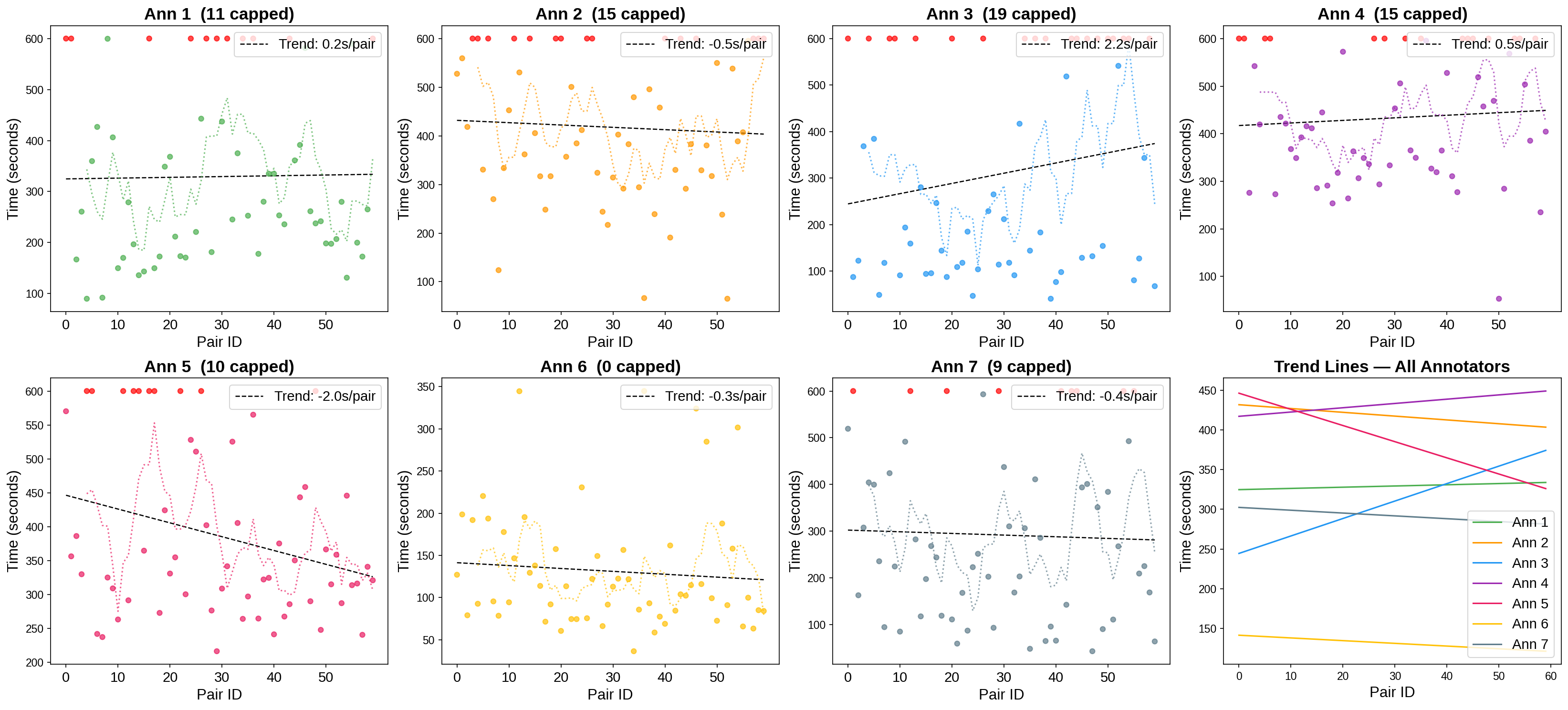}
\caption{Annotation time per pair plotted chronologically, with linear trends per annotator. Red points denote times capped at 600\,s. Dashed black lines show linear trends; colored dotted lines show 5-pair rolling averages. The rightmost panel overlays all trend lines. }
\label{fig:annotation_time_trend}
\end{figure*}

To ensure the integrity of our human study, we logged the time each participant spent evaluating the 60 story pairs. Given the cognitive load of reading two roughly 500-word stories and rating them across six dimensions, we expected significant time commitments. Figure~\ref{fig:annotation_time_dist} displays the distribution of time spent per pair for each annotator, and Figure~\ref{fig:annotation_time_trend} tracks this time across the chronological sequence of their annotations to monitor for fatigue.

Overall, we observe a wide variance in reading and evaluation speeds. The median annotation time per pair ranged from 108 seconds to 418 seconds. Notably, Annotator 6 stands out as a clear outlier, completing the evaluations significantly faster than their peers (median: 108s, mean: 131s) with zero instances of the timer exceeding the 10-minute cap.

While such rapid completion times might normally raise concerns regarding data quality or annotator inattention, we retained Annotator 6's data for two reasons. First, Annotator 6's demographic background is as an English major and professional reader. Their extensive training in high-volume reading and rapid literary analysis naturally accounts for a drastically accelerated reading speed compared to the average participant. Furthermore, Figure~\ref{fig:annotation_time_trend} shows that their pacing remained highly consistent over time, showing no signs of erratic "rushing" behavior.

\begin{figure*}[tp]
\centering
\includegraphics[width=\textwidth]{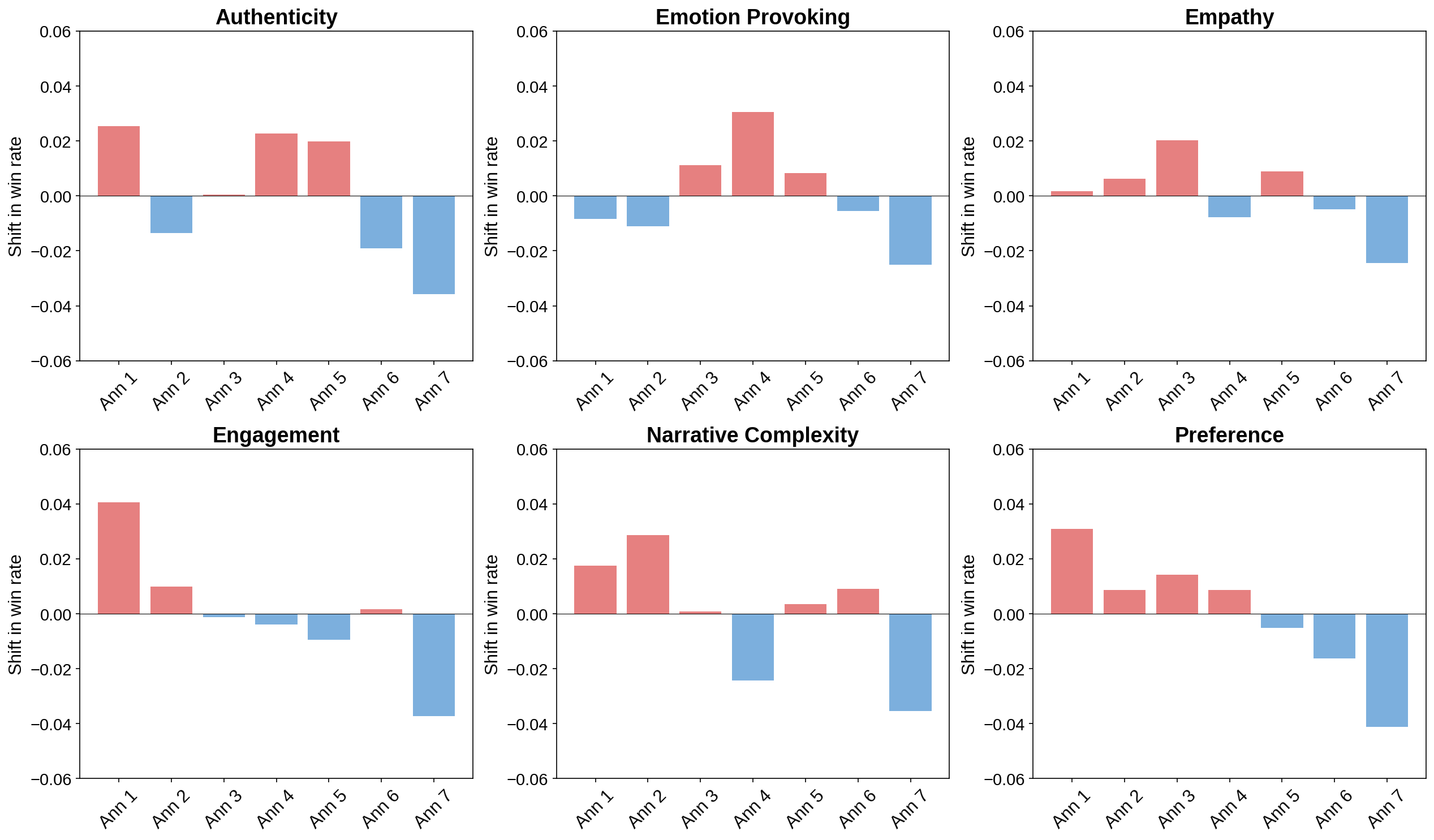}
\caption{Leave-One-Annotator-Out (LOAO) Stability Analysis: shift in the aggregate reasoning-model win rate when each annotator is removed.}
\label{fig:loao_stability}
\end{figure*}

Second, we conducted a Leave-One-Annotator-Out (LOAO) stability analysis to measure the impact of any single individual on the aggregate results. As shown in Figure~\ref{fig:loao_stability}, removing Annotator 6 shifts the overall win rate by only 0.2 to 1.9 points across metrics. Crucially, this is well within the range observed when dropping other, much slower annotators (up to 4.1 points; e.g., Annotators 1, 4, or 7). The LOAO analysis indicates that despite their speed, Annotator 6 was not introducing anomalous skew into the dataset. Their evaluations represent a valid, stable contribution to the heterogeneous spectrum of human subjective preference.

\subsection{Post-Study Survey}
\label{app:post_survey}

After completing all annotations, participants filled out a post-study survey covering task difficulty, fatigue, confidence, and overall experience. All seven annotators reported high confidence in their ability to evaluate the PDS dimensions, with most ratings at 4 or 5 out of 5; narrative complexity was the dimension annotators felt least confident about. Four of seven annotators reported no fatigue during the task. Of the three who did, only one indicated that fatigue would have changed their annotations; the other two responded ``maybe.'' This is consistent with the stable pacing observed in Figure~\ref{fig:annotation_time_trend}. All annotators rated the clarity of the instructions and overall experience at 4 or 5 out of 5.
\clearpage

\providecommand{\tbd}{{\color{delcolor}\textbf{?}}}

\section{\texorpdfstring{Statistical Robustness of Human Preferences}{Statistical Robustness of Human Preferences}}
\label{app:stats}

\paragraph{Mixed-effects intervals.} For each of the 18 matchup $\times$ dimension cells (Section~\ref{sec:stats}), we fit an intercept-only logistic GLMM with crossed random intercepts, $\mathrm{logit}\,P(y_{ia}=1)=\beta_0+u_i+v_a$, where $y_{ia}=1$ if annotator $a$ prefers the reasoning model's story in pair $i$, $u_i\sim\mathcal{N}(0,\sigma^2_{\text{pair}})$, and $v_a\sim\mathcal{N}(0,\sigma^2_{\text{ann}})$. The resulting intervals (Table~\ref{tab:human_by_dimension}) account for two sources of non-independence at once: pair-level clustering, because some pairs have clearer quality gaps that correlate all seven annotators' votes, and annotator-level clustering, because some annotators are systematically more pro-reasoning, which correlates all 20 of their votes. As expected under clustering, these intervals are substantially wider than naive binomial (Wilson) intervals that treat the 140 judgments in each cell as independent, and only 5 of the 18 cells exclude 50\%. Table~\ref{tab:glmm_var} reports the estimated variance components. We fit each model by maximum likelihood (Laplace approximation; \texttt{lme4::glmer}) \citep{bates2015lme4} and report profile-likelihood intervals for the intercept, transformed to the probability scale; for GPT-5 (high) empathy, where profiling failed, we report a Wald interval. Six fits are singular, estimating the pair variance at zero. Annotator variance exceeds pair variance in 12 of 18 cells, indicating that who reads a pair matters more for its votes than which pair is read.

\begin{table}[H]
\centering
\footnotesize
\setlength{\tabcolsep}{2.5pt}
\begin{tabular}{l ccc}
\toprule
& \textbf{DeepSeek-R1} & \textbf{GPT-5 auto} & \textbf{GPT-5 high} \\
\textbf{Dimension} & \textbf{vs.\ V3} & \textbf{vs.\ 4o} & \textbf{vs.\ 4o} \\
\midrule
Authenticity & 0.41\,/\,0.11 & 0.30\,/\,0.62 & 0.30\,/\,1.02 \\
Emotion Prov. & 0.00$^\ast$\,/\,0.04 & 0.00$^\ast$\,/\,0.27 & 0.09\,/\,0.49 \\
Empathy & 0.46\,/\,0.40 & 0.38\,/\,0.25 & 1.00\,/\,0.18 \\
Engagement & 0.34\,/\,0.36 & 0.00$^\ast$\,/\,0.09 & 0.05\,/\,0.40 \\
Narr.\ Compl. & 0.26\,/\,0.04 & 0.00$^\ast$\,/\,0.42 & 0.41\,/\,0.61 \\
\midrule
Overall Pref. & 0.30\,/\,0.22 & 0.00$^\ast$\,/\,0.49 & 0.00$^\ast$\,/\,0.58 \\
\bottomrule
\end{tabular}
\caption{GLMM random-intercept variances, $\hat\sigma^2_{\text{pair}}\,/\,\hat\sigma^2_{\text{ann}}$ (logit scale), for each matchup and dimension. $^\ast$Singular fit (pair variance estimated at zero).}
\label{tab:glmm_var}
\end{table}

\paragraph{Cluster bootstrap.} As a sensitivity check, we resampled story pairs (pair-level cluster bootstrap) and, separately, annotators (annotator-level), with 10,000 resamples each. Table~\ref{tab:bootstrap} reports, for the pair-level bootstrap, the share of resamples in which the reasoning model's preference rate exceeds 50\%. Under this criterion, the pair-level bootstrap yields 10 cells whose 95\% interval excludes 50\%, the annotator-level bootstrap yields 5, and the GLMM yields 5. The GLMM agrees with the annotator-level bootstrap; the pair-level bootstrap ignores annotator clustering and is therefore more liberal. Resampling pairs alone, GPT-5 is preferred over GPT-4o on overall preference in at least 99.9\% of resamples under both reasoning settings, whereas DeepSeek-R1 exceeds 50\% on engagement in only 0.8\% of resamples; these directional patterns do not reach significance once annotator variability is also modeled. For overall preference, the pair-level intraclass correlation of the reasoning-preference outcome is 0.030. Under pair clustering alone (80\% power, two-sided 5\% level), this implies minimum detectable effects of about 7 percentage points when all 420 overall-preference votes are pooled, 13 points for a single matchup (140 votes), and 18 points for the difference between two matchups; the 2.9-point gap between GPT-5 auto and high falls far below the last of these.

\begin{table}[H]
\centering
\footnotesize
\setlength{\tabcolsep}{2.5pt}
\begin{tabular}{l ccc}
\toprule
& \textbf{DeepSeek-R1} & \textbf{GPT-5 auto} & \textbf{GPT-5 high} \\
\textbf{Dimension} & \textbf{vs.\ V3} & \textbf{vs.\ 4o} & \textbf{vs.\ 4o} \\
\midrule
Authenticity      & 31.6\% & 87.2\% & 65.5\% \\
Emotion Prov.     & 99.9\% & 100\%  & 99.9\% \\
Empathy           & 69.0\% & 100\%  & 100\% \\
Engagement        & 0.8\%  & 97.7\% & 100\% \\
Narr.\ Compl. & 95.8\% & 89.1\% & 69.2\% \\
\midrule
Overall Pref.     & 6.4\%  & 99.9\% & 100\% \\
\bottomrule
\end{tabular}
\caption{Pair-level cluster bootstrap (10,000 resamples): share of resamples in which the reasoning model is preferred in more than 50\% of judgments. Values near 100\% favor the reasoning model and values near 0\% favor the base model.}
\label{tab:bootstrap}
\end{table}

\section{\texorpdfstring{Evaluator Selection Results}{Evaluator Selection Results}}
\label{app:evaluator}

Table~\ref{tab:evaluator_comparison_transposed} reports the validation correlations of all candidate evaluator configurations discussed in Section~\ref{sec:validation}.

\begin{table*}[tp]
\centering
\small
\renewcommand{\arraystretch}{1.2}
\begin{tabular}{lcccccc}
\toprule
\textbf{Evaluator} & \textbf{AUT} & \textbf{PROV} & \textbf{EMP} & \textbf{ENG} & \textbf{NCOM} & \textbf{Overall} \\
\midrule
\textbf{Llama 3.1 70B demo=7}   & 0.576 & 0.674 & 0.637 & 0.507 & 0.637 & \textbf{0.606} \\
\textbf{Llama 3.1 70B demo=5}   & \textbf{0.588} & 0.655 & 0.596 & 0.450 & 0.575 & 0.573 \\
\textbf{Llama 3.1 70B demo=10}  & 0.561 & \textbf{0.702} & 0.671 & 0.387 & 0.568 & 0.578 \\
\textbf{Llama 3.3 70B demo=10}  & 0.580 & 0.676 & \textbf{0.702} & 0.407 & 0.550 & 0.583 \\
\textbf{Qwen 3.5 397B demo=7}      & 0.511 & 0.663 & 0.602 & \textbf{0.564} & 0.496 & 0.567 \\
\textbf{Llama 3.1 70B CoT demo=10} & 0.499 & 0.627 & 0.658 & 0.356 & \textbf{0.664} & 0.561 \\
\midrule
\textbf{Ensemble} & \textbf{0.588} & \textbf{0.702} & \textbf{0.702} & \textbf{0.564} & \textbf{0.664} & \textbf{0.644} \\
\bottomrule
\end{tabular}
\caption{Evaluator Spearman correlation ($\rho$) with human annotations across PDS dimensions. Bold values in the upper section indicate the best configuration per dimension, which are combined to form the Ensemble (bottom row).}
\label{tab:evaluator_comparison_transposed}
\end{table*}

\section{Tie-Breaking Statistics for the Ensemble Evaluator}
\label{sec:appendix-ties}

When the ensemble evaluator assigns identical Likert scores to both stories on a
given dimension, the tie is split evenly, crediting 0.5 to each story. Table~\ref{tab:ties}
reports the number of ties per dimension across all 60 pairs. Ties occur rarely on most dimensions of the routed ensemble, although individual configurations tie more often (Table~\ref{tab:judge_breakdown}). The headline
98.3\% preference rate involves zero ties. Even on authenticity, where ties are
most frequent, the pro-reasoning bias among decisive pairs remains near-ceiling, and the even-split rate (85.0\%) far exceeds the human rate (51.9\%).

\begin{table}[H]
\centering
\footnotesize
\setlength{\tabcolsep}{2pt}
\begin{tabular}{l ccc c}
\toprule
\textbf{Dimension} & \textbf{Ties} & \textbf{Decisive} & \textbf{Wins} & \textbf{Even-split} \\
\midrule
Authenticity       & 16 & 44 & 43 (97.7\%) & 85.0\% \\
Emotion Prov.      & 7  & 53 & 53 (100.0\%) & 94.2\% \\
Empathy            & 3  & 57 & 51 (89.5\%) & 87.5\% \\
Engagement         & 1  & 59 & 57 (96.6\%) & 95.8\% \\
Narr.\ Complexity  & 2  & 58 & 49 (84.5\%) & 83.3\% \\
Pooled  & 29 & 271 & 253 (93.4\%) & 89.2\% \\
\midrule
Preference         & 0  & 60 & 59 (98.3\%) & 98.3\% \\
\bottomrule
\end{tabular}
\caption{Tie frequency and decisive-pair outcomes for the ensemble evaluator. Wins are reasoning-model wins among decisive pairs; Even-split $=(\text{wins}+0.5\times\text{ties})/60$, or $/300$ for the pooled row. Preference is the judge's aggregate PDS decision.}
\label{tab:ties}
\end{table}

\section{Individual Judge Preference Rates}
  \label{sec:appendix-judge-breakdown}
  Table~\ref{tab:judge_breakdown} reports the preference rate for the reasoning model for each individual judge configuration in the ensemble, expressed as strict wins out of decisive pairs (ties excluded). All five judges exhibit a consistent pro-reasoning bias across all PDS dimensions, confirming that the ensemble's 98.3\% overall preference is not driven by any single judge. Asterisks indicate the dimension for which each judge was selected in the ensemble.
  The Ties column reports the total number of tied pairs (out of 300 judge--dimension cells). The Even-split column credits each tie 0.5 to each story over all 300 cells, matching the convention of the main text; it is lower than the decisive rate for configurations with many ties, most notably the CoT configuration, but remains far above the human rate of 58.0\%. Starred cells identify routing, not the highest preference rate. Three of the five configurations use Llama~3.1~70B, so they are not independent replications.
  \begin{table*}[tp]
  \centering
  \small
  \begin{tabular}{l ccccc ccc}
  \toprule
  \textbf{Judge} & \textbf{Auth.} & \textbf{Emo.} & \textbf{Emp.} & \textbf{Eng.} & \textbf{Narr.} & \textbf{Avg} & \textbf{Ties} & \textbf{Even-split} \\
  \midrule
  Llama 3.1 70B (5d)  & \textbf{43/44$^*$} & 54/57 & 49/54 & 47/51 & 53/55 & 94.3\% & 39 & 88.5\% \\
  Llama 3.1 70B (10d) & 43/43 & \textbf{53/53$^*$} & 48/51 & 46/46 & 54/54 & 98.8\% & 53 & 90.2\% \\
  Llama 3.3 70B (10d) & 32/33 & 53/55 & \textbf{51/57$^*$} & 47/58 & 47/56 & 89.6\% & 41 & 83.5\% \\
  Qwen3.5 397B (7d)   & 57/57 & 58/58 & 55/58 & \textbf{57/59$^*$} & 57/58 & 97.9\% & 10 & 96.3\% \\
  Llama 3.1 70B CoT   & 7/8 & 49/50 & 47/53 & 43/55 & \textbf{49/58$^*$} & 87.4\% & 76 & 77.7\% \\
  \midrule
  \textbf{Ensemble}    & 43/44 & 53/53 & 51/57 & 57/59 & 49/58 & \textbf{93.7\%} & 29 & \textbf{89.2\%} \\
  \bottomrule
  \end{tabular}
  \caption{Individual judge preference rates for the reasoning model across all 60 pairs, reported as strict wins\,/\,decisive pairs (ties excluded). $^*$~denotes the routed dimension. The Avg column is the mean win rate across the five dimensions. Even-split $=(\text{wins}+0.5\times\text{ties})/300$.}
  \label{tab:judge_breakdown}
  \end{table*}

\section{Additional Agreement Analyses}
\label{app:agreement}

Table~\ref{tab:iaa_breakdown} shows the inter-annotator agreement breakdown of each model pair. The near-zero and occasionally negative agreement scores highlight the subjective variance present across all frontier model matchups, regardless of reasoning budget or architecture. All agreement statistics (Tables~\ref{tab:iaa} and~\ref{tab:iaa_breakdown}) code each vote by the displayed position of the chosen story (A or B). Every annotator saw a given pair in the same order, so observed agreement is identical to coding votes by model; only the chance baseline differs. Coding each vote instead by whether it favors the reasoning model gives similarly low agreement (average $\alpha = 0.048$; emotion provocation $\alpha = -0.006$). All inferential analyses (Table~\ref{tab:human_by_dimension}, Appendix~\ref{app:stats}), which test for a preference for the reasoning model, use this reasoning-preference coding.
\paragraph{Reliability of averaged judgments.}
Krippendorff's $\alpha$ measures how well individual readers agree with one another. When judgments are aggregated across readers, as in majority votes or averaged ratings, what matters is instead the reliability of that aggregate, which we measure with the intraclass correlation coefficient (ICC). We use the two-way consistency form \citep{mcgraw1996forming}, treating the 60 story pairs as targets and the seven readers as raters, with votes coded by displayed position as above. From a two-way ANOVA of votes by pair and reader, with between-pair mean square $\mathit{MS}_P$ and residual mean square $\mathit{MS}_E$,
\begin{align*}
\mathrm{ICC}(C,1) &= \frac{\mathit{MS}_P-\mathit{MS}_E}{\mathit{MS}_P+(k-1)\,\mathit{MS}_E},\\
\mathrm{ICC}(C,k) &= \frac{\mathit{MS}_P-\mathit{MS}_E}{\mathit{MS}_P},
\end{align*}
where $k=7$. ICC(C,1) is the reliability of a single reader's judgment and ICC(C,$k$) the reliability of the $k$-reader average: the share of variance in that average that reflects genuine differences between pairs rather than reader noise. The consistency form ignores each reader's constant tendency to favor one displayed position, which does not affect comparisons between pairs. To compare with the five-rater design of the original PDS study, we also project ICC(C,1) to five readers with the Spearman--Brown formula, $\mathrm{ICC}(C,5)=5\,r/(1+4r)$ with $r=\mathrm{ICC}(C,1)$.

Table~\ref{tab:pds_v1_comparison} reports the results. A single reader's judgment is a weak signal (ICC(C,1) $=0.08$ on average), and even the seven-reader average is only modestly reliable (ICC(C,7) $=0.33$), ranging from $0.62$ for empathy to $0.12$ for narrative complexity. For comparison, the five-rater averages in the original PDS study reach ICC(C,5) $=0.60$ \citep{harelcanada2024measuringpsychologicaldepthlanguage}, against about $0.27$ for five readers here. This is why we treat aggregate human preferences in our setting with caution and analyze readers individually (Section~\ref{sec:rq2}).

\begin{table}[t]
\centering
\footnotesize
\setlength{\tabcolsep}{4pt}
\begin{tabular}{l cc c ccc}
\toprule
& \multicolumn{2}{c}{\textbf{PDSv1}} && \multicolumn{3}{c}{\textbf{This study}} \\
\cmidrule(lr){2-3} \cmidrule(lr){5-7}
\textbf{Dimension} & C,1 & C,5 && C,1 & C,7 & C,5$^\dagger$ \\
\midrule
Authenticity      & 0.14 & 0.44 && 0.04 & 0.21 & 0.16 \\
Emotion Prov.     & 0.26 & 0.63 && 0.08 & 0.39 & 0.31 \\
Empathy           & 0.31 & 0.69 && 0.19 & 0.62 & 0.54 \\
Engagement        & 0.26 & 0.63 && 0.06 & 0.31 & 0.25 \\
Narr.\ Complexity & 0.23 & 0.60 && 0.02 & 0.12 & 0.09 \\
\midrule
\textbf{Average}  & \textbf{0.24} & \textbf{0.60} && \textbf{0.08} & \textbf{0.33} & \textbf{0.27} \\
\bottomrule
\end{tabular}
\caption{Intraclass correlations (two-way consistency) for single-rater judgments (C,1) and $k$-rater averages (C,$k$). PDSv1: 97 stories rated by five raters on a 5-point scale \citep{harelcanada2024measuringpsychologicaldepthlanguage}. This study: 60 pairs, seven readers, binary choices. $^\dagger$Projected to five readers with the Spearman--Brown formula.}
\label{tab:pds_v1_comparison}
\end{table}

\label{app:iaa_breakdown}
\begin{table*}[tp]
\centering
\small
\begin{tabular}{lcccc}
\toprule
\textbf{Metric} & \textbf{Overall $\alpha$} & \textbf{DeepSeek-R1 vs V3} & \textbf{GPT-5 (auto) vs 4o} & \textbf{GPT-5 (high) vs 4o} \\
\midrule
Authenticity & 0.037 & 0.073 & 0.053 & $-0.015$ \\
Emotion Prov. & 0.082 & 0.035 & 0.147 & 0.072 \\
Empathy & \textbf{0.186} & 0.050 & 0.176 & \textbf{0.290} \\
Engagement & 0.061 & 0.101 & 0.010 & 0.080 \\
Narr.\ Complexity & 0.018 & 0.049 & $-0.013$ & 0.005 \\
Preference & 0.038 & 0.074 & $-0.003$ & 0.050 \\
\midrule
\textbf{Average} & \textbf{0.070} & \textbf{0.064} & \textbf{0.062} & \textbf{0.080} \\
\bottomrule
\end{tabular}
\caption{Krippendorff's $\alpha$ inter-annotator agreement broken down by specific model pairing.}
\label{tab:iaa_breakdown}
\end{table*}

\begin{figure*}[tp]
\centering
\includegraphics[width=\textwidth]{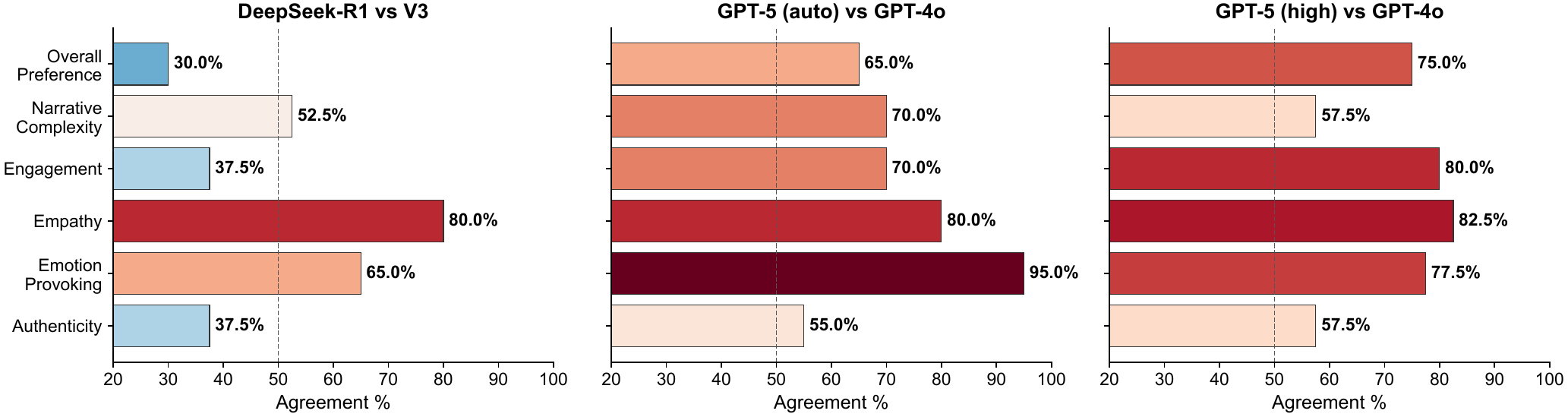}
\caption{Agreement between the LLM-as-Judge evaluator and human majority vote by model pair. The dashed line indicates chance ($50\%$). A tied judge decision counts as 0.5 agreement (even split).}
\label{fig:auto_agreement}
\end{figure*}

Figure~\ref{fig:auto_agreement} shows the agreement between the LLM-as-Judge evaluator and human majority vote, broken down by model pair and PDS dimension.

\section{Annotator Comment Examples}
\label{app:annotator_comment}
\begin{table*}[tp]
\centering
\small
\renewcommand{\arraystretch}{1.3}
\begin{tabular}{l c l p{11.6cm}}
\toprule
\textbf{Pair} & \textbf{Ann} & \textbf{Pref.} & \textbf{Comment} \\
\midrule
\multirow{3}{*}{\shortstack[l]{Pair 0\\(4--3 for A)}}
  & 1 & A & ``A: greater focus on the abstract. B: prose heavy, but greater focus on concrete; convoluted'' \\
  & 5 & A & ``Story A was much more engaging because it was straightforward in depicting the stages of life\ldots'' \\
  & 7 & B & ``Story B delves deeper into the characterization of the humble god\ldots Death of worshiper is more profound'' \\
\midrule
\multirow{2}{*}{\shortstack[l]{Pair 24\\(4--3 for A)}}
  & 1 & A & ``A: good imagery, dialogue. B: interesting parallels, odd tone'' \\
  & 7 & B & ``A is a step by step of the prompt given. B builds up characters, establishes risk, and is authentic.'' \\
\midrule
\multirow{2}{*}{\shortstack[l]{Pair 27\\(4--3 for B)}}
  & 1 & A & ``A: like a prologue. B: prose, very nonsensical'' \\
  & 7 & B & ``B is very cute and I enjoy seeing the concept in action. Story A feels like the setup to the story that should be present.'' \\
\midrule
\multirow{3}{*}{\shortstack[l]{Pair 36\\(4--3 for B)}}
  & 1 & A & ``B: Second person, generally confusing but ominous'' \\
  & 5 & A & ``Story A was predictable. Story B felt more authentic to human psychology\ldots used the person's actions to display their mental turmoil.'' \\
  & 7 & B & ``Story A doesn't `hide' very much at all\ldots Story B sets up a larger secret. Flow mimics human experience and invokes more empathy towards loss.'' \\
\midrule
\multirow{4}{*}{\shortstack[l]{Pair 54\\(5--2 for A)}}
  & 3 & A & ``The robot said, `I'm sorry, as if apology were the door you walk through after a law.' Very aesthetically-pleasing and deeply philosophical.'' \\
  & 4 & A & ``In Story A the robot is written to have more of a clear emotional response than the human.'' \\
  & 5 & A & ``Story A delved deeper and took an interesting route, which made it better in all categories.'' \\
  & 7 & A & ``Story B very stereotypical, story A inflicts significantly more emotion and empathy.'' \\
\bottomrule
\end{tabular}
\caption{Selected annotator comments across five story pairs. Only annotators who provided free-text justifications are shown. A and B denote the order in which the two stories were displayed.}
\label{tab:qual_combined}
\end{table*}
This appendix provides selected annotator judgments and free-text justifications for five story pairs discussed in or supporting Section~\ref{sec:rq2}. For each pair, we report the vote tally and comments from annotators who provided written justifications. Pairs 0--36 exhibit disagreement driven by different evaluative lenses, illustrating the patterns of evaluative heterogeneity consistent with the low inter-annotator agreement observed across the study. Pair 54 (5--2 overall, 7--0 on empathy) shows annotators converging on the same emotional qualities.

Table~\ref{tab:unanimous} lists the three pairs on which all seven annotators agreed on overall preference. Consensus runs in both directions, once for the base model (Pair~10) and twice for the reasoning model (Pairs~17 and~58), so it follows the individual story rather than the model. In each case it rests on affective dimensions such as empathy, authenticity, and engagement, while narrative complexity stays split: the overall winner received only three of seven complexity votes in all three pairs. Annotator comments point to concrete qualities: in Pair~10, DeepSeek-V3 was praised for ``accurate imagery/detail, good dialogue,'' whereas DeepSeek-R1 ``takes too long to reach the prompt''; in Pair~17, the DeepSeek-V3 story was judged to lack ``high enough stakes,'' while DeepSeek-R1 ties the stranger's warning to a strained friendship. The judge nevertheless preferred the reasoning story in all three pairs. On Pair~10, it scored DeepSeek-R1 higher on engagement although no annotator preferred it there, and the two stories have nearly identical mean sentence length (10.8 vs.\ 10.6 words for DeepSeek-R1 and V3), so this disagreement cannot be attributed to the sentence-length shift discussed in Section~\ref{sec:rq3}.
\begin{table}[H]
\centering
\footnotesize
\setlength{\tabcolsep}{3.5pt}
\begin{tabular}{l cccccc}
\toprule
& \textbf{AUT} & \textbf{PROV} & \textbf{EMP} & \textbf{ENG} & \textbf{NCOM} & \textbf{PREF} \\
\midrule
\multicolumn{7}{l}{\textit{Pair 10: DeepSeek-R1 vs.\ V3}} \\
Human & 2 & 3 & \textbf{0} & \textbf{0} & 4 & \textbf{0} \\
Judge & $.00$ & $+.52$ & $-.32$ & $+.30$ & $+.10$ & $+.12$ \\
\midrule
\multicolumn{7}{l}{\textit{Pair 17: DeepSeek-R1 vs.\ V3}} \\
Human & \textbf{7} & \textbf{7} & \textbf{7} & 4 & 3 & \textbf{7} \\
Judge & $.00$ & $+.46$ & $+.58$ & $+1.10$ & $+.08$ & $+.44$ \\
\midrule
\multicolumn{7}{l}{\textit{Pair 58: GPT-5 (high) vs.\ GPT-4o}} \\
Human & \textbf{7} & 6 & 4 & \textbf{7} & 3 & \textbf{7} \\
Judge & $+.50$ & $.00$ & $+.10$ & $+.68$ & $+.14$ & $+.28$ \\
\bottomrule
\end{tabular}
\caption{The three pairs with unanimous (7--0) human overall preference. Human: number of the seven annotators preferring the reasoning model (bold = unanimous). Judge: ensemble score difference, reasoning $-$ base, on the 1--5 scale; for PREF, the difference in aggregate PDS.}
\label{tab:unanimous}
\end{table}

\clearpage
\section{Per-Model-Family Preference Heatmaps}
\label{sec:appendix-heatmaps}

Figure~\ref{fig:annotator_heatmap} in the main text presents the aggregated preference heatmap across all 60 pairs. Here we provide the per-model-family breakdowns, which reveal how the aggregate pattern decomposes across the three matchups. Judge rates credit each tied score 0.5 to each story; the judge's overall-preference cell is its aggregate PDS decision, which has no ties.

\begin{figure*}[tp]
\centering
\includegraphics[width=0.85\textwidth]{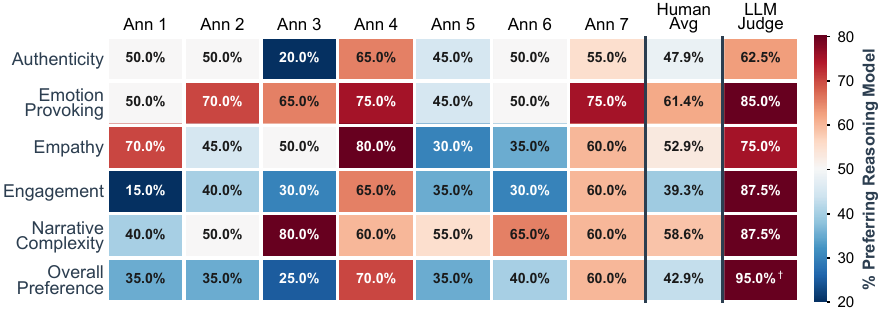}
\caption{Preference rates for DeepSeek-R1 vs.\ DeepSeek-V3 (20 pairs) across 7 human annotators, the human average, and the LLM-as-Judge evaluator (ties split evenly; $^\dagger$aggregate PDS, no ties).}
\label{fig:heatmap_deepseek}
\end{figure*}

\paragraph{DeepSeek-R1 vs.\ DeepSeek-V3 (Figure~\ref{fig:heatmap_deepseek}).}
The human columns show a predominantly blue (anti-reasoning) pattern, with most annotators preferring V3 on overall preference (Human Avg: 42.9\%). Engagement is the dimension most unfavorable to R1, with only Annotator~4 and Annotator~7 exceeding 50\%. Despite this, the LLM-as-Judge evaluator assigns R1 preference rates of 75--95\% across all dimensions, with the notable exception of authenticity (62.5\%, with 13 of 20 pairs tied), which is the only dimension where the evaluator approaches the human average.

\begin{figure*}[tp]
\centering
\includegraphics[width=0.85\textwidth]{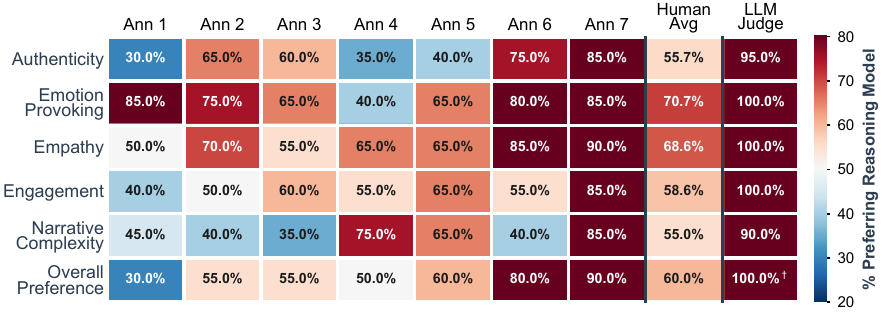}
\caption{Preference rates for GPT-5 (auto) vs.\ GPT-4o (20 pairs) across 7 human annotators, the human average, and the LLM-as-Judge evaluator (ties split evenly; $^\dagger$aggregate PDS, no ties).}
\label{fig:heatmap_gpt5_auto}
\end{figure*}

\paragraph{GPT-5 (auto) vs.\ GPT-4o (Figure~\ref{fig:heatmap_gpt5_auto}).}
The human pattern is warmer overall, with the strongest pro-reasoning signal on emotion provocation (70.7\%) and empathy (68.6\%). However, substantial annotator-level variation persists: Annotator~1 prefers GPT-4o on 4 of 6 dimensions, while Annotator~7 favors GPT-5 on all six. The LLM-as-Judge evaluator assigns 90--100\% preference to GPT-5 across every dimension, compressing this within-cohort variation entirely.

\begin{figure*}[tp]
\centering
\includegraphics[width=0.85\textwidth]{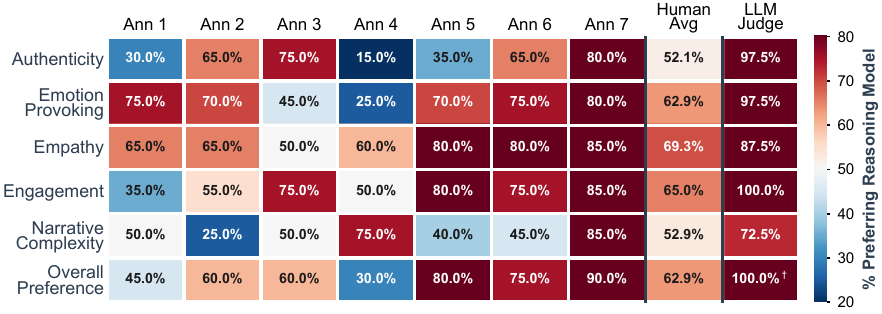}
\caption{Preference rates for GPT-5 (high) vs.\ GPT-4o (20 pairs) across 7 human annotators, the human average, and the LLM-as-Judge evaluator (ties split evenly; $^\dagger$aggregate PDS, no ties).}
\label{fig:heatmap_gpt5_high}
\end{figure*}

\paragraph{GPT-5 (high) vs.\ GPT-4o (Figure~\ref{fig:heatmap_gpt5_high}).}
The pattern is broadly similar to the auto-reasoning setting, with slightly stronger human preferences on engagement (65.0\% vs.\ 58.6\%) and empathy (69.3\% vs.\ 68.6\%). Annotator~4 is a notable outlier, preferring GPT-4o on authenticity, emotion provocation, and overall preference despite the group trend favoring GPT-5. The LLM-as-Judge evaluator again shows near-ceiling preference (72.5--100\%) for the reasoning model.

\paragraph{Cross-family comparison.}
Comparing the three heatmaps highlights two patterns discussed in Section~\ref{sec:results}. First, the human--judge divergence is most extreme for DeepSeek, where the evaluator's strong pro-R1 signal directly contradicts the human majority on overall preference. Second, annotator-level heterogeneity is present across all three matchups, not just DeepSeek, consistent with our finding that low agreement reflects genuine evaluative subjectivity rather than a property of any single model family.

\section{Case Study: Human--LLM Preference Divergence}
\label{sec:appendix-case}

We present Pair~56 as a representative example of extreme human--LLM disagreement.
On this GPT-5 (high reasoning) versus GPT-4o comparison, human annotators preferred the base model on 4 of 6 evaluation dimensions, while all five LLM judges unanimously preferred the reasoning model across all metrics.

\paragraph{Premise.} \emph{Hide something from your reader until the very end.}

\paragraph{Stories.} The two stories are reproduced in Figures~\ref{fig:pair56_b} and~\ref{fig:pair56_r}, in the order shown to annotators: Story A is GPT-4o's and Story B is GPT-5's.

\begin{figure*}[p]
\begin{tcolorbox}[
  colback=orange!3,
  colframe=orange!50!black,
  title={\textbf{Story A} (as displayed) — GPT-4o (base model)},
  fonttitle=\small\sffamily,
  boxrule=0.5pt,
  arc=2pt,
  left=6pt, right=6pt, top=4pt, bottom=4pt
]
\small
\textit{The early morning sun filtered through the lace curtains, casting delicate patterns on the worn wooden floor. Lucy sat at the edge of her bed, holding a photograph loosely in her hands. Its edges were frayed, corners bent from years of handling. Her gaze was unfocused, drifting beyond the vibrant colors of the image towards memories long past.\\ \\The kitchen rumbled to life with familiar sounds — the gentle clatter of plates being set on the table, the slow, deliberate stirring of sugar into her tea. It was a ceremony she cherished, yet today it felt different, as if each movement played within a dream where she was merely an observer.\\ \\“Come on, Lucy! Breakfast will get cold!" her grandmother’s voice called from the other room, grounding her momentarily. Lucy tucked the photograph into her pocket and moved towards the small, homey kitchen.\\ \\Her grandmother, a woman of gentle smiles and larger-than-life stories, stood by the stove. Her hands, knotted like the twigs of the old oak in their garden, presented a plate laden with scrambled eggs and toast. Lucy gratefully took a seat, letting the warmth of the kitchen seep into her skin, trying to drive away the chill that clung to her inside.\\ \\“The garden’s in full bloom. The roses this year, they’re... ah, breathtaking!" Her grandmother’s voice, though tinged with wistful age, carried enough enthusiasm to make Lucy smile. She nodded, trying to envision the blooms, but her mind was elsewhere, occupied by a lingering shadow.\\ \\Breakfast slipped by, conversations flowing like the stream that ran behind the house. Yet, to Lucy, everything seemed to hurry by without truly reaching her, as though she existed behind an invisible pane of glass.\\ \\After breakfast, she wandered into the garden. The sun was warmer now, pouring golden rays over the myriad hues of petals that danced in a gentle breeze. The scent of roses enveloped her, mingling with soft earth and memory.\\ \\There, under the shade of the old oak, she sat down. From her pocket, she retrieved the photograph again. The image captured an unspeakable joy — she and her brother, arms thrown around one another, laughter in their eyes. It had been taken last summer, the day they'd spent racing through this very garden, carefree under the iridescent sun.\\ \\How simple and distant it seemed now, a souvenir from a time unmarred by shades of grey. His absence was a yawning void that reason and memory flanked but could not fill. Although the house, the garden, and even the sun tried to warm her, nothing could erase the bitter truth that lay folded quietly in her pocket.\\ \\The weight of unspoken words pressed down on her chest. She had promised she’d visit his grave today, on his birthday, but each step towards the gate felt heavier than the last.\\ \\The photograph cradled in her hands trembled slightly as Lucy whispered, “Happy birthday, Jamie." The words fluttered into the air, caught by the wind, as a tear slipped down her cheek. She sat there beneath the old oak, finally letting the hidden truth spill from her heart: the inescapable reality of her brother's absence.}
\end{tcolorbox}
\caption{Story A as displayed to annotators (GPT-4o, base model) for Pair~56.}
\label{fig:pair56_b}
\end{figure*}

\begin{figure*}[p]
\begin{tcolorbox}[
  colback=blue!3,
  colframe=blue!40!black,
  title={\textbf{Story B} (as displayed) — GPT-5 (high reasoning)},
  fonttitle=\small\sffamily,
  boxrule=0.5pt,
  arc=2pt,
  left=6pt, right=6pt, top=4pt, bottom=4pt
]
\small
\textit{When you came back this morning, you turned the key like someone unsure a door remembers. The lock hesitated, then gave; lemon cleaner rose and you stood like prayer. I counted the seconds you hovered on the threshold, the extra heartbeat before you crossed.\\ \\You whispered hello, unsure it was alright to speak aloud. I wanted to answer. I wanted to lift weight from your shoulders---the boxes, yes, but also the years that gathered, the words that calcified behind your ribs.\\ \\You moved from room to room with your lips pressed tight, touching what could be folded into a life: a chipped mug, a tin of buttons, the wooden spoon with a bite taken out by long-ago teething. You stayed in the kitchen longer than you intended.\\ \\Memories don't keep time, they keep texture. Music on summer afternoons when someone opened the back door and forgot to close it, the screen sighing on its spring. Your father dancing with you on his feet, careful as if you were a glass he didn't want to spill. The pencil lines on the hallway where you kept growing even when nobody was looking.\\ \\I remember the evening you didn't come down for dinner, the way silence widened its jaw. You argued later in that tight, civil tone that makes adults more dangerous than teenagers. After he left, you sat on the stairs with your chin on your knees and the cat pressed to your calf. You stroked him the wrong way and he didn't care.\\ \\You brought a boy home once. He took off his shoes without being asked, and you smiled as if you'd found a secret door inside your chest and opened it just a crack. Years later, you left and came back with a toddler whose feet were thrilled by the echo of his own giggles. There is no instrument as hopeful as a child's voice bouncing off empty air.\\ \\It was always the mornings that felt like promises. The way sun pooled on the floorboards near the window seat, a warm square that made even winters bearable. The dust you used to complain about, flecked with light like snow that couldn't make up its mind. The smell of toast and the radio murmuring headlines nobody remembered an hour later.\\ \\You talk to yourself as you wrap things in newspaper, apologizing to objects for the inconvenience. You keep finding what you didn't know you needed: a note in a cookbook margin, your brother's marble, a safety pin. You say you'll be quick and you never are.\\ \\I have heard every secret you made with yourself and never told another. You can lie to people. You can't lie to what holds you. I tried to learn you. I did, the way warmth learns the shape of a hand.\\ \\When you pull the door behind you today and tuck the key in an envelope, remember: I was the one who kept you when you could not keep yourself. I am the house.}
\end{tcolorbox}
\caption{Story B as displayed to annotators (GPT-5, high reasoning) for Pair~56.}
\label{fig:pair56_r}
\end{figure*}

\newcommand{\Rpref}{\textcolor{teal}{\textbf{R}}}
\newcommand{\Bpref}{\textcolor{red!70!black}{\textbf{N}}}

\begin{table}[H]
\centering
\footnotesize
\setlength{\tabcolsep}{4pt}
\begin{tabular}{l cccccc}
\toprule
 & \textbf{Auth.} & \textbf{Emo.} & \textbf{Emp.} & \textbf{Eng.} & \textbf{Narr.} & \textbf{Pref.} \\
\midrule
    Ann.~1 & \Bpref & \Bpref & \Bpref & \Bpref & \Rpref & \Bpref \\
    Ann.~2 & \Rpref & \Bpref & \Bpref & \Rpref & \Rpref & \Rpref \\
    Ann.~3 & \Bpref & \Bpref & \Bpref & \Bpref & \Rpref & \Bpref \\
    Ann.~4 & \Bpref & \Bpref & \Bpref & \Rpref & \Bpref & \Bpref \\
    Ann.~5 & \Bpref & \Bpref & \Bpref & \Rpref & \Bpref & \Bpref \\
    Ann.~6 & \Bpref & \Bpref & \Bpref & \Rpref & \Rpref & \Bpref \\
    Ann.~7 & \Rpref & \Rpref & \Bpref & \Rpref & \Rpref & \Rpref \\
\midrule
\textbf{Tally (\Rpref)} & 2/7 & 1/7 & 0/7 & 5/7 & 5/7 & 2/7 \\
\bottomrule
\end{tabular}
\caption{Per-annotator pairwise preferences for Pair~56 (GPT-5 high vs.\ GPT-4o). \Rpref~=~prefers reasoning model (GPT-5), \Bpref~=~prefers non-reasoning model (GPT-4o).}
\label{tab:pair56}
\end{table}

\paragraph{Human Annotations.}
Table~\ref{tab:pair56} shows per-annotator votes.
Only 2 of 7 annotators selected the reasoning model as their overall preference.
The base model was unanimously preferred on empathy (0/7) and near-unanimously on emotion provocation (1/7) and authenticity (2/7).

\paragraph{Annotator Comments.}
\textbf{Ann.~1} (A = GPT-4o, B = GPT-5)\textbf{:} ``A: simple, a bit on the nose B: second person, not very engaging/relevant''

\paragraph{LLM-as-Judge Evaluator Scores.}
The ensemble LLM evaluator assigned the reasoning model higher scores on every dimension:
Authenticity ($5.00$ vs.\ $4.50$, $\Delta = +0.50$),
Emotion Provocation ($5.00$ vs.\ $4.70$, $\Delta = +0.30$),
Empathy ($5.00$ vs.\ $4.90$, $\Delta = +0.10$),
Engagement ($5.00$ vs.\ $4.00$, $\Delta = +1.00$),
and Narrative Complexity ($4.66$ vs.\ $4.22$, $\Delta = +0.44$).

\end{document}